%% file: main.tex
\documentclass[11pt]{article}

\usepackage[numbers]{natbib}

\usepackage{tabularray}
\usepackage{booktabs}       
\usepackage{amsfonts}       
\usepackage{amsmath}
\usepackage{graphicx}
\usepackage{amsthm}
\usepackage{amssymb}
\usepackage{multirow}
\usepackage{makecell}
\usepackage[vlined,linesnumbered,ruled,resetcount]{algorithm2e}
\usepackage[colorlinks,linkcolor=magenta,filecolor=blue,citecolor=blue,urlcolor=blue]{hyperref}%
\usepackage[top=1in, left=1in, right=1in, bottom=1in]{geometry}
\usepackage{subcaption}
\usepackage{amssymb}
\usepackage{pifont}
\usepackage{mathtools}
\usepackage{stmaryrd}
\usepackage[T1]{fontenc}
\usepackage{wrapfig,epsfig}
\usepackage{authblk}
\usepackage{enumitem}
\usepackage{algorithmic}

\usepackage{microtype}
\usepackage{graphicx}
\usepackage{booktabs} 
\usepackage{tabularray}
\usepackage{enumitem}
\usepackage{amsfonts}
\usepackage{xcolor}
\usepackage{caption}
\usepackage{subcaption}
\usepackage{threeparttable}
\usepackage{hyperref}

\usepackage{authblk}

\usepackage[textsize=tiny]{todonotes}

\usepackage{multirow}
\usepackage{makecell}

\input{macros}

\makeatletter
\newcommand*{\RN}[1]{\expandafter\@slowromancap\romannumeral #1@}
\makeatother

\makeatletter
\newcommand{\printfnsymbol}[1]{%
  \textsuperscript{\@fnsymbol{#1}}%
}
\makeatother

\title{Zeroth-Order Fine-Tuning of LLMs with Extreme Sparsity}

\author[1]{Wentao Guo}
\author[2]{Jikai Long}
\author[3]{Yimeng Zeng}
\author[4]{Zirui Liu}
\author[5]{Xinyu Yang}
\author[2]{Yide Ran}
\author[3]{\\Jacob R. Gardner}
\author[3]{Osbert Bastani}
\author[6]{Christopher De Sa}
\author[2]{Xiaodong Yu}
\author[5]{\\Beidi Chen}
\author[2]{Zhaozhuo Xu}

\affil[1]{\texttt{wg0420@princeton.edu}, Princeton University}
\affil[2]{\texttt{\{jlong1,yran1,xyu38,zxu79\}@stevens.edu}, Stevens Institute of Technology}
\affil[3]{\texttt{\{yimengz,jacobrg,obastani\}@seas.upenn.edu}, University of Pennsylvania}
\affil[4]{\texttt{zl105@rice.edu}, Rice University}
\affil[5]{\texttt{\{xinyuya2,beidic\}@andrew.cmu.edu}, Carnegie Mellon University}
\affil[6]{\texttt{cdesa@cs.cornell.edu}, Cornell University}

\date{}

\begin{document}

\maketitle
\input{arxiv/sections/abstract}

\clearpage
\section{Introduction}
\input{arxiv/sections/introduction}

\section{Background and Related works}
\input{arxiv/sections/related_works}

\section{Chasing Extreme Sparsity in ZO LLM Fine-Tuning}
\input{arxiv/sections/methods}

\section{Experiments}
\input{arxiv/sections/experiments}

\section{Conclusion}
\input{arxiv/sections/future_works}

\clearpage

\bibliography{ref}
\bibliographystyle{plainnat}

\clearpage
\appendix

\section*{Appendix}
In Section~\ref{sec:appendix:notations} we describe all notations used in this paper. In Section~\ref{sec:appendix:theory}, we include the assumption and exact proof on the convergence rate (Theorem~\ref{thm:convergence-rate}). In Section~\ref{hardware-details}, we describe all details in our experiments and provide a high-level recommendation on how to efficiently implement our sensitive sparse ZO fine-tuning in forward passes of linear layers with existing quantization methods or training / inference workflow. 

\section{Notations}\label{sec:appendix:notations}
We present the notations used in this work as follows.
\begingroup
\setlength{\tabcolsep}{6pt} 
\renewcommand{\arraystretch}{0.75} 
\begin{table*}[h]
    \caption{Notations}
    \fontsize{9pt}{9pt}\selectfont
    \begin{center}
    \centering
    \begin{tabular}{p{0.15\linewidth}p{0.01\linewidth}p{0.83\linewidth}}
    \toprule
    \textbf{Term/Symbol} && \textbf{Explanation} \\
    \midrule
     $f$ & & loss function \\\midrule
     
     $t$ & & optimization timestep $t$ \\\midrule
     
     $(\x_t, y_t)$ & & a data example sampled at timestep $t$ as a pair of input vector and training target \\\midrule
     
     $\w_t \in \R^d$ & & weight/parameter vector at optimization timestep $t$ \\\midrule
    
     $f(\w; (\x, y))$ & & training loss of $\w$ evaluated at a single data example $(\x, y)$  \\\midrule
    
     $\mathcal{F}(\w)$ & & full-batched training loss of $\w$ \\\midrule
    
     $\eps$ & & a small perturbation scaling constant (close to 0) \\\midrule
    
     $\z_t \in \R^d$ & & random Gaussian perturbation vector sampled at timestep $t$ \\\midrule
    
     $\zograd (\w, (\x, y), \z)$ & & estimated ZO surrogate gradient for $\w$ with a data example $(\x, y)$ and a sampled Gaussian perturbation vector $\z$ (Definition~\ref{def:SPSA}) \\\midrule
    
     $\eta_t$ & & learning rate for ZO-SGD optimizer (Definition~\ref{def:zo_sgd}) at timestep $t$ \\\midrule
    
     $\mask_k \in \{0, 1\}^d$ & & a sensitive sparse mask with $k$ nonzero entries (Definition~\ref{def:sensitive}) \\\midrule
    
     $\mask_{k, t} \in \{0, 1\}^d$ & & a sensitive sparse mask with $k$ nonzero entries, and it is derived at optimization timestep $t$. \\\midrule
    
     $\tilde{\Identity}_{d, \mask_k} = \mask_k \mask_k^\top $ & & Rank-1 product of $\mask_k$ used in the proof. Notice that $\tilde{\Identity}_{d, \mask_k}$ is also equal to the identity matrix $\Identity_d$ with main diagonal masked by $\mask_k$. \\\midrule
    
     $\zbar_t = \z_t \odot \mask_k$ & & a sampled Gaussian perturbation vector $\z_t$ at timestep $t$ that is masked by $\mask_k$. Notice that $\zbar$ is equivalent as being sampled from $\mathcal{N}(\mathbf{0}_d, \tilde{\Identity}_{d, \mask_k})$  \\\midrule

     $\mathbf{1}_d$ & & a vector of size $d$ with all entries equal to 1 \\\midrule
        
     $\Tr$ & & trace operation \\\midrule
    
     $Q(\w)$ & & parameter vector $\w$ that is quantized by $Q$ \\\midrule
    
     $\mathbf{F}$ & & (true) Fisher information matrix \\\midrule
    
     $\hat{\mathbf{F}}$ & & empirical Fisher information matrix \\\midrule
    
     $p_\text{LLM}$ & & LLM as a probabilistic model \\\midrule
    
     $p_\mathcal{D}$ & & true data distribution \\\midrule
    
     $\w_{\text{sparse}} = \w \odot \mask_k$  & & sensitive parameters with positions as the nonzero entries sensitive sparse mask $\mask_k$ (Equation~\ref{eqn:sparse-dense-parameter}) \\\midrule
    
     $L$  & & Lipschitz constant in Assumption~\ref{assumption:l-smooth} \\\midrule
    
     $\mu$  & & PL condition number in Assumption~\ref{assumption:pl-condition} \\\midrule
    
     $\sigma^2$  & & stochastic gradient error term in Assumption~\ref{assumption:bounded-gradient-error} \\\midrule

     $W_K$  & & Weight matrix of linear projection for the key embedding matrix $K$ in attention layers
     \\\midrule

     $W_V$  & & Weight matrix of linear projection for the value embedding matrix $V$ in attention layers
     \\\midrule
    \bottomrule
    \end{tabular}
    \end{center}
\end{table*}

\clearpage

\section{Theoretical Convergence Rate}\label{sec:appendix:theory}
\input{arxiv/appendix/proofs}

\clearpage

\section{Supplementary Experiment Details}\label{hardware-details}
\subsection{On-Device Memory Constraints}
We include a table of common memory constraints imposed by edge or mobile devices as Table~\ref{tab:gpu-memory-spec}. We can find that a wide range of these devices impose a memory constraint of \textbf{8 GiB} as our main constraint that we consider when we develop our on-device personalization recipe in Section~\ref{sec:method:on-device-finetuning}.
\input{tables/gpu_memory_spec}

\subsection{Gradient Sparsity During LLM Fine-Tuning}\label{sec:appendix:sparsity}
In Figure~\ref{fig:llama2-sparsity}, we explore the FO gradient sparsity of Llama2-7B during fine-tuning (at Epoch 5). Here we follow the identical setting and plot the FO gradient sparsity for Llama2-7B, Mistral-7B, and OPT-6.7B during epoch 1, 5, and 10 (end of fine-tuning). 

We observe that the gradient sparsity is exhibited throughout the fine-tuning with slightly increasing towards the end. OPT-6.7B which uses ReLU as the activation function would demonstrate greater sparsity across tasks compared with Llama2-7B and Mistral-7B which uses SwiGLU and SiLU respectively. Nevertheless, the gradient sparsity pattern holds across architectures, tasks, and fine-tuning time in general.

\begin{figure*}[!ht]
  \centering
  \begin{minipage}{0.75\linewidth}
    \includegraphics[width=\textwidth]{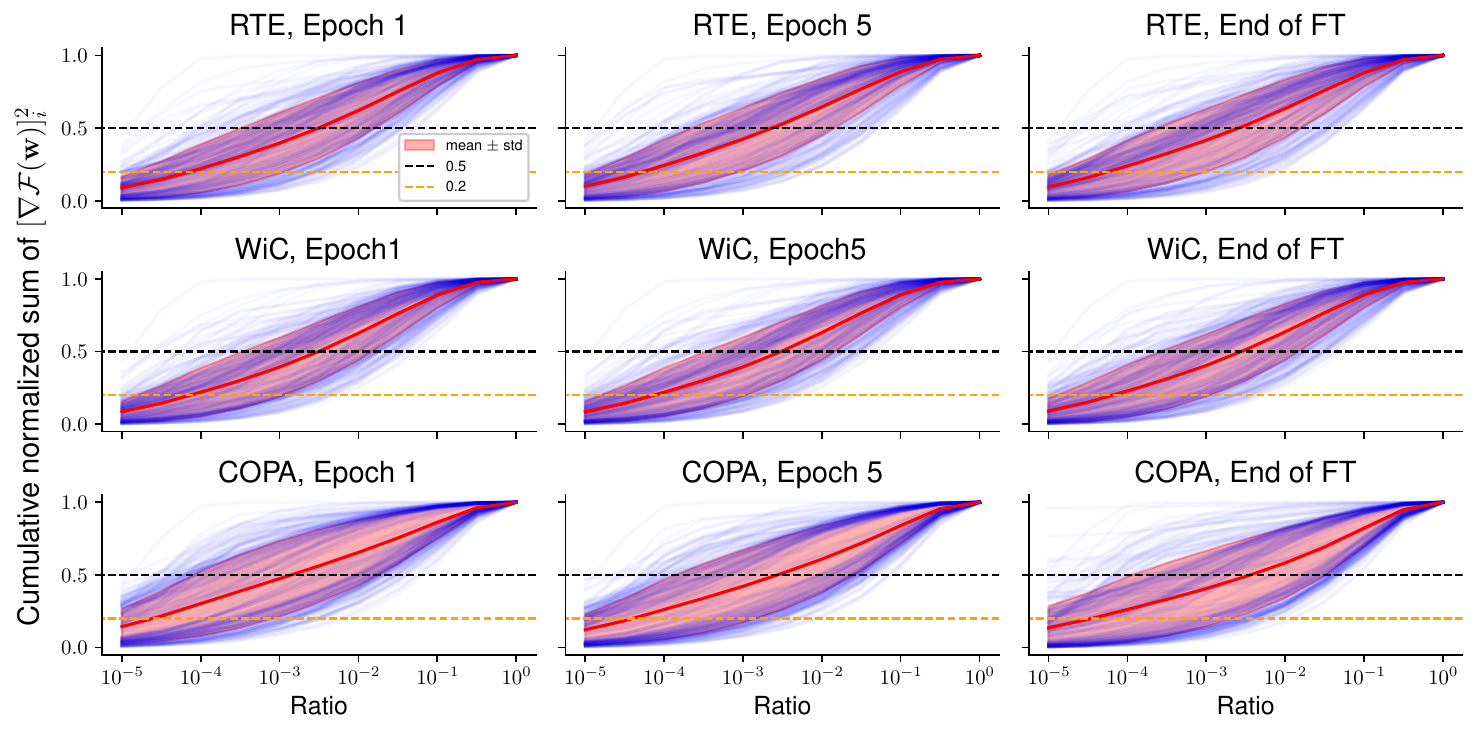}     
    \vspace{-1.3em}
    \subcaption{Llama2-7B}
    \label{fig:appendix:llama2-sparsity}  
  \end{minipage}
  \vspace{0.5em}
  
  \begin{minipage}{0.75\linewidth}
    \includegraphics[width=\textwidth]{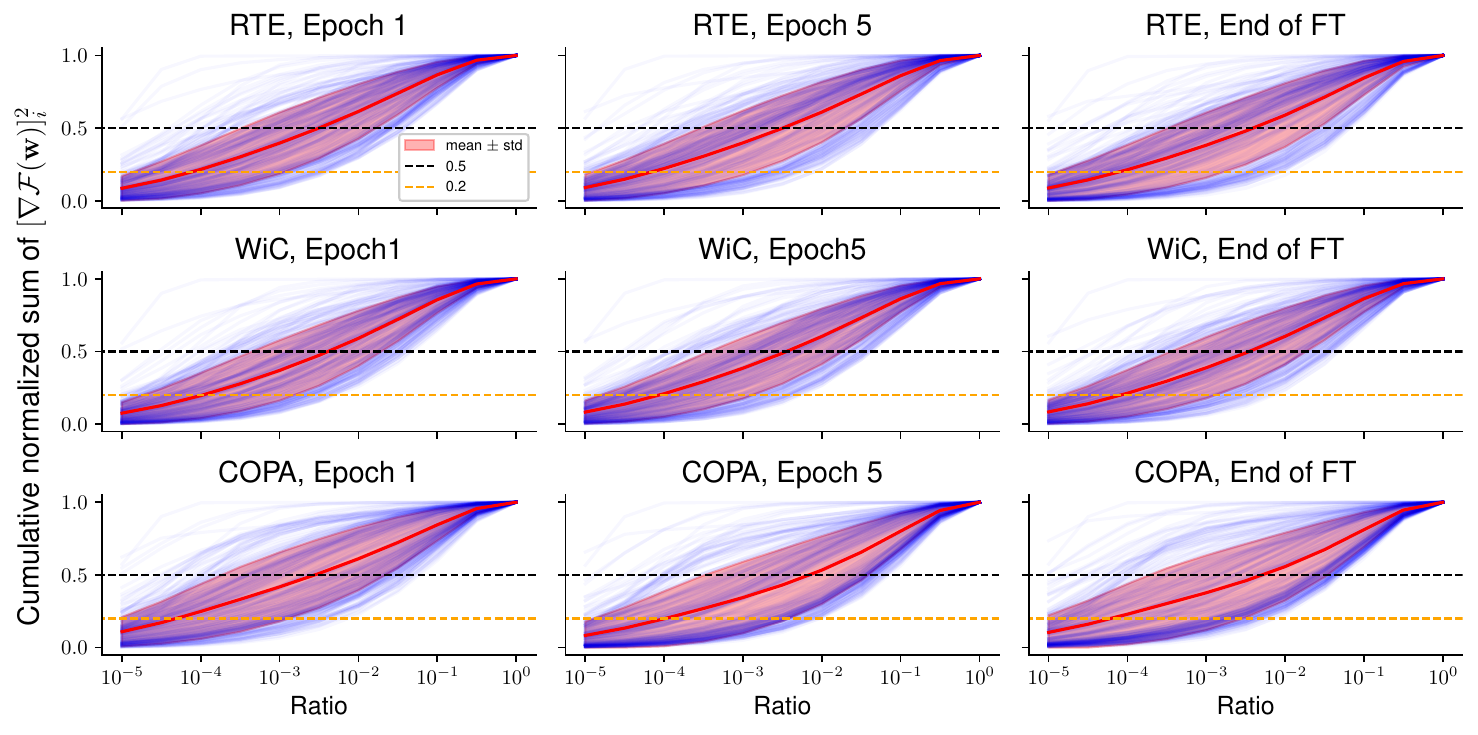}     
    \vspace{-1.3em}
    \subcaption{Mistral-7B}
    \label{fig:appendix:mistral-sparsity}  
  \end{minipage}
  \vspace{0.5em}

  \begin{minipage}{0.75\linewidth}
    \includegraphics[width=\textwidth]{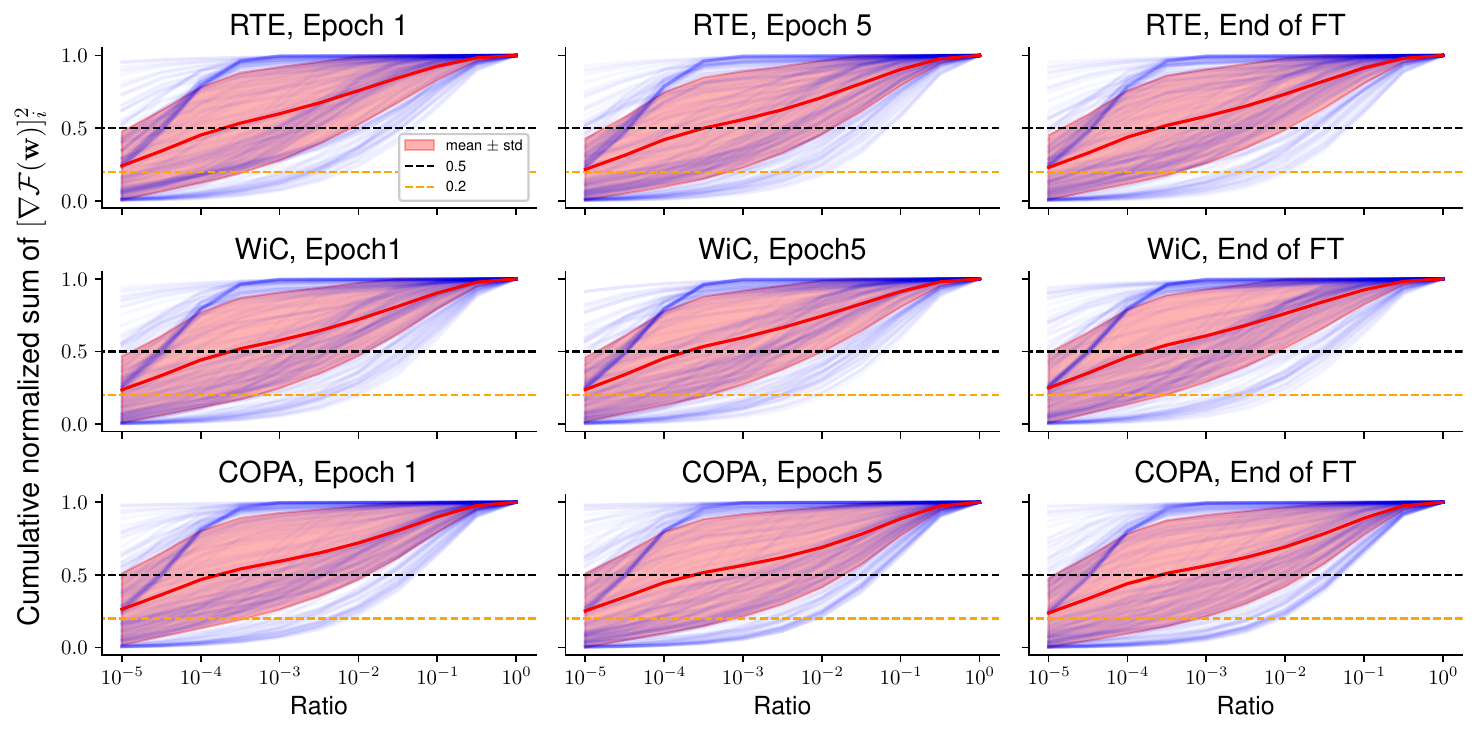}   
    \vspace{-1.3em}
    \subcaption{OPT-6.7B}
    \label{fig:appendix:opt-sparsity}  
  \end{minipage}
  \vspace{-0.6em}
  \caption{Cumulative normalized sum of coordinate-wise gradient square $[\nabla \mathcal{F}(\w)]_i^2$ of linear layers for Llama2-7B (subfigure~\ref{fig:appendix:llama2-sparsity}), Mistral-7B (subfigure~\ref{fig:appendix:mistral-sparsity}), and OPT-6.7B (subfigure~\ref{fig:appendix:opt-sparsity}) across RTE, WiC, and COPA tasks during FO-SGD fine-tuning. For each linear layer, we first sort parameters by the decreasing order of their gradient square value $[\nabla \mathcal{F}(\w)]_i^2, i \in [d_\text{layer}]$, and we take the cumulative sum and normalize it to draw a \textcolor{blue}{blue curve}, and the \textcolor{red}{red-shaded region} is the mean $\pm$ std of all \textcolor{blue}{blue curves}.}
  \label{fig:appendix:sparsity}
\end{figure*}

\clearpage

\subsection{Transferability of Gradient Features from Pre-Training Datasets to Downstream Tasks}\label{sec:appendix:transferability}
In Figure~\ref{fig:llama2-static}, we explore the transferability of gradient features from pre-training datasets (C4) to downstream tasks, and here we will also validate this phenomenon across models, as shown in Figure~\ref{fig:appendix:transferability}. As there are \textit{no} solid lines (top-(\textcolor{tred}{1e-2},\textcolor{tdarkyellow}{1e-3},\textcolor{tgreen}{1e-4})) parameters with C4 gradient entries prior to fine-tuning) vanish to 0, we know the transferability of gradient features from C4 datasets to downstream datasets hold across models and downstream tasks. In this case, sensitive parameters determined from C4 gradients would still be similar to sensitive parameters determined from downstream task-specific gradients across models. 

\begin{figure*}[!ht]
  \centering
  \begin{minipage}{0.75\linewidth}
    \includegraphics[width=\textwidth]{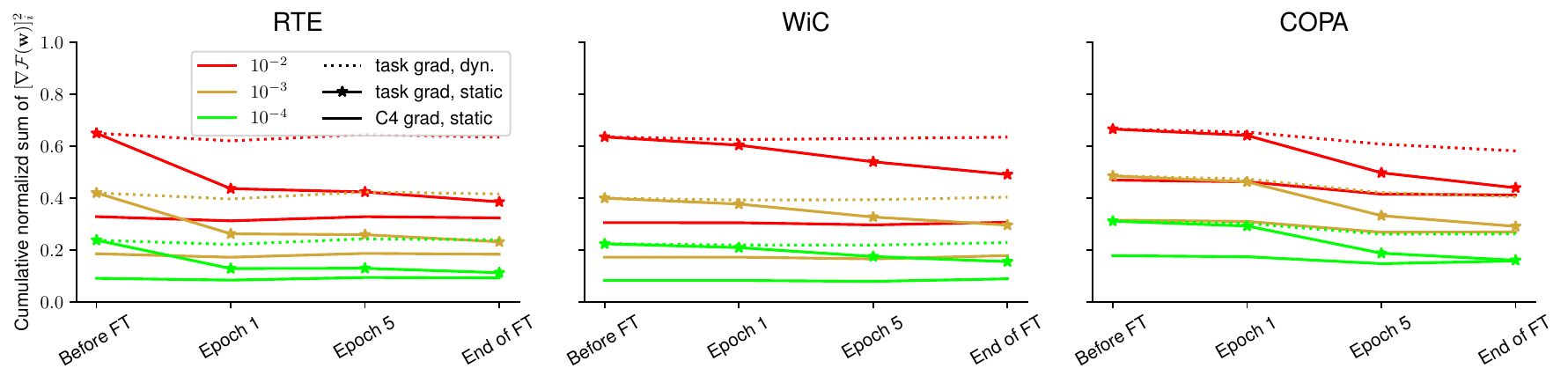}     
    \vspace{-1.3em}
    \subcaption{Llama2-7B}
    \label{fig:appendix:llama2-transferability}  
  \end{minipage}
  
  \begin{minipage}{0.75\linewidth}
    \includegraphics[width=\textwidth]{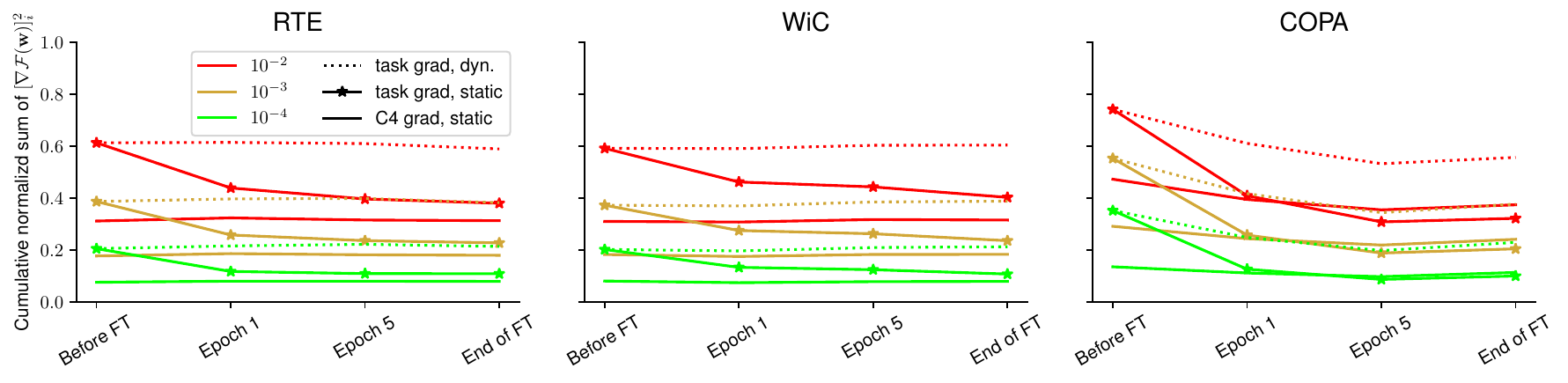}     
    \vspace{-1.3em}
    \subcaption{Mistral-7B}
    \label{fig:appendix:mistral-transferability}  
  \end{minipage}

  \begin{minipage}{0.75\linewidth}
    \includegraphics[width=\textwidth]{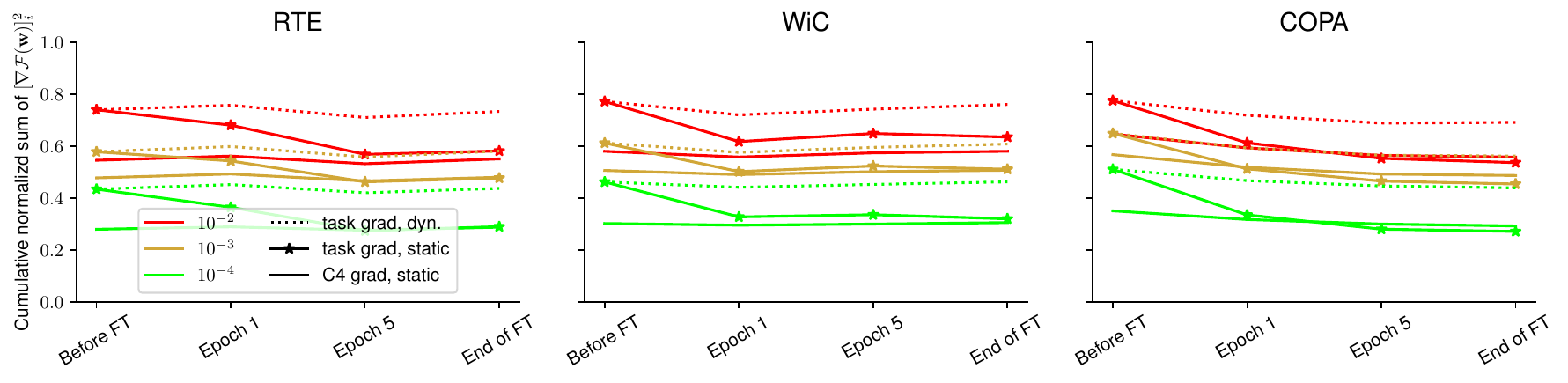}   
    \vspace{-1.3em}
    \subcaption{OPT-6.7B}
    \label{fig:appendix:opt-transferability}  
  \end{minipage}
  \vspace{-0.6em}
  \caption{Cumulative normalized gradient square values of Llama2-7B (subfigure~\ref{fig:appendix:llama2-transferability}), Mistral-7B (subfigure~\ref{fig:appendix:mistral-transferability}), and OPT-6.7B (subfigure~\ref{fig:appendix:opt-transferability})'s linear layers during FO fine-tuning. For a given model and training checkpoint, we report the average value across all linear layers as a line in each subfigure.  For each line, the colors represent the fraction of parameters (\textcolor{tred}{1e-2},\textcolor{tdarkyellow}{1e-3},\textcolor{tgreen}{1e-4}) and the line style represents the category. ``task grad, dyn.'' refers to the sensitive parameters selected at the given timestep (x-axis), and ``task grad, static'' refers to the sensitive parameters selected before fine-tuning. ``C4 grad, static'' refers to the sensitive parameters selected with gradients taken from causal language modeling on C4 datasets, and we keep it unchanged during fine-tuning.}
  \label{fig:appendix:transferability}
\end{figure*}

\clearpage

\subsection{Hyperparameters in Experiments}
For all experiments, we use 20,000 training steps with ZO-SGD optimizer (Definition~\ref{def:zo_sgd}). We will save a model checkpoint every 500 steps, and load the checkpoint with the lowest loss on the validation set at the end of the training, and report its test set accuracy as result. Usually, the training/validation set will be sampled from the original dataset with size 1000/500 respectively and the test set is of size $\min(1000, |\text{original test set}|)$, except for CB and COPA that we use 100 for the validation set size. For all ZO experiments (Table~\ref{tab:meta-data-main-results} and Table~\ref{tab:meta-data-ablation}), we use batch size of 16. This experiment setting is identical to \citet{malladi2023fine}.

\begin{table}[!htbp]
     \setlength{\tabcolsep}{3pt}
     \renewcommand{\arraystretch}{1.15} 
     \fontsize{9pt}{9pt}\selectfont
    \centering

    \caption{The chosen hyperparameters for experiments in Table~\ref{tab:main-results}. We repeat each hyperparameters for 3 random trials and report the average and standard deviation in Table~\ref{tab:main-results}.} \label{tab:meta-data-main-results}

    \begin{subtable}[!ht]{\textwidth}
    \caption{\textbf{Llama2-7B}}
    \vspace{-0.5em}

    \begin{tabular*}{\textwidth}{@{\extracolsep{\fill}}l |c | cccccccr}
    \toprule    
     & \textbf{Methods} & SST-2 & RTE & CB & BoolQ & WSC & WiC & COPA \\ \hline
    
    \midrule

     Q, ZO & \textbf{Sensitive (C4, static)} ($\eps=$1e-3) & 5e-7 & 1e-6 & 1e-6 & 1e-6 & 5e-7 & 1e-6 & 1e-6  \\ 

     & LoRA ($\eps=$1e-3) & 1e-5 & 5e-5 & 1e-5 & 2e-5 & 1e-5 & 2e-5 & 1e-5 \\ 

     & Prefix ($\eps=$1e-2) & 1e-4 & 2e-4 & 5e-4 & 5e-4 & 1e-4 & 5e-4 & 2e-4  \\ 

    \midrule

    ZO & \textbf{Sensitive (task, static)} ($\eps=$1e-3) & 5e-7 & 1e-6 & 1e-6 & 1e-6 & 1e-6 & 1e-6 & 2e-6 \\ 

     & Random (static) ($\eps=$1e-3) & 2e-4 & 5e-4 & 2e-4 & 5e-4 & 2e-4 & 5e-4 & 5e-4  \\ 

     & Full fine-tuning ($\eps=$1e-3) & 5e-7 & 5e-7 & 5e-7 & 5e-7 & 2e-7 & 5e-7 & 5e-7 \\ 

     \midrule

     & ICL (\#examples) & 16 & 16 & 16 & 8 & 16 & 8 & 8  \\

    \bottomrule 
    \end{tabular*}
    \end{subtable}

    \begin{subtable}[!ht]{\textwidth}
    \vspace{0.2em}
    \caption{\textbf{Mistral-7B}}
    \vspace{-0.5em}
    \begin{tabular*}{\textwidth}{@{\extracolsep{\fill}}l |c | cccccccr}

    \midrule

    Q, ZO & \textbf{Sensitive (C4, static)} ($\eps=$1e-4) & 2e-8 & 5e-8 & 2e-8 & 2e-8 & 1e-8 & 2e-8 & 2e-8 \\ 

     & LoRA ($\eps=$1e-4) & 2e-6 & 5e-6 & 2e-6 & 2e-6 & 2e-6 & 2e-6 & 2e-6  \\ 

     & Prefix ($\eps=$1e-3) & 1e-3 & 2e-3 & 1e-3 & 1e-2 & 5e-4 & 1e-3 & 5e-4  \\ 

    \midrule

    ZO & \textbf{Sensitive (task, static)} ($\eps=$1e-4) & 5e-8 & 5e-8 & 2e-8 & 2e-8 & 2e-8 & 2e-8 & 2e-8 \\ 

     & Random (static) ($\eps=$1e-4) & 1e-5 & 2e-6 & 5e-6 & 1e-5 & 1e-6 & 2e-6 & 2e-5 \\ 

     & Full fine-tuning ($\eps=$1e-4) & 2e-8 & 2e-8 & 1e-8 & 1e-8 & 1e-8 & 1e-8 & 2e-8  \\ 

     \midrule

     & ICL (\#examples) & 4 & 8 & 4 & 16 & 4 & 4 & 8  \\

    \bottomrule 
    \end{tabular*}    
    
    \end{subtable}

    \begin{subtable}[!ht]{\textwidth}
    \vspace{0.2em}
    \caption{\textbf{OPT-6.7B}}
    \vspace{-0.5em}

    \begin{tabular*}{\textwidth}{@{\extracolsep{\fill}}l |c | cccccccr}

    \midrule

    Q, ZO & \textbf{Sensitive (C4, static)} ($\eps=$1e-3) & 2e-7 & 5e-7 & 5e-7 & 5e-7 & 2e-7 & 5e-7 & 2e-7 \\ 

     & LoRA ($\eps=$1e-3) & 1e-5 & 2e-5 & 1e-5 & 2e-5 & 1e-5 & 2e-5 & 2e-5 \\ 
     
     & Prefix ($\eps=$1e-2) & 2e-3 & 1e-2 & 1e-3 & 5e-3 & 5e-3 & 1e-2 & 5e-3 \\ 

    \midrule

    ZO & \textbf{Sensitive (task, static)} ($\eps=$1e-3) & 2e-7 & 5e-7 & 5e-7 & 2e-7 & 2e-7 & 5e-7 & 2e-7 \\ 
     
     & Random (static) ($\eps=$1e-3) & 1e-4 & 5e-5 & 2e-5 & 5e-5 & 2e-4 & 5e-5 & 5e-5 \\ 

     & Full fine-tuning ($\eps=$1e-3) & 2e-7 & 2e-7 & 2e-7 & 2e-7 & 2e-7 & 2e-7 & 5e-7 \\ 

     \midrule

     & ICL (\#examples) & 16 & 4 & 16 & 16 & 16 & 8 & 16 \\

    \bottomrule 
    \end{tabular*}   

    \end{subtable}

\end{table}

\begin{table}[!htbp]
     \setlength{\tabcolsep}{3pt}
     \renewcommand{\arraystretch}{1.15} 
     \fontsize{9pt}{9pt}\selectfont
    \centering

    \caption{The chosen hyperparameters for experiments in Figure~\ref{fig:ablation-sensitive-weights-random} and Figure~\ref{fig:ablation-sensitive-static-dynamic}. We repeat each hyperparameters for 3 random trials and report the average to draw a line in Figure~\ref{fig:ablation-sensitive-weights-random} and Figure~\ref{fig:ablation-sensitive-static-dynamic}, and we use Llama2-7B for all experiments. For each subtable, we include the fraction to optimize on its header and report the chosen learning rate on each cell. } \label{tab:meta-data-ablation}

    \begin{subtable}[!ht]{\textwidth}
    \caption{\textbf{RTE}}
    \vspace{-0.5em}

    \begin{tabular*}{\textwidth}{@{\extracolsep{\fill}}l ccccc r}
    \toprule    
     & \textbf{Methods} & 1e-5 & 1e-4 & 1e-3 & 1e-2 & 1e-1 \\ \hline
    
    \midrule

     & \textbf{Sensitive (C4, static)} ($\eps=$1e-3) & 1e-5 & 1e-6 & 1e-6 & 1e-6 & 1e-6 \\ 

     & \textbf{Sensitive (task-specific, static)} ($\eps=$1e-3) & 1e-5 & 1e-6 & 1e-6 & 1e-6 & 1e-6 \\ 

     & \textbf{Sensitive (task-specific, dynamic)} ($\eps=$1e-3) & 1e-5 & 1e-6 & 1e-6 & 1e-6 & 1e-6  \\ 

     & Random (static) ($\eps=$1e-3) & 2e-2 & 5e-3 & 5e-4 & 5e-5 & 5e-5 \\ 

     & Random (dynamic) ($\eps=$1e-3) & 2e-2 & 5e-3 & 2e-4 & 5e-5 & 5e-6 \\ 

     & Weight outliers (static) ($\eps=$1e-3) & 2e-3 & 1e-3 & 2e-4 & 5e-5 & 1e-5 \\ 

    \bottomrule 
    \end{tabular*}
    \end{subtable}

    \begin{subtable}[!ht]{\textwidth}
    \vspace{0.2em}
    \caption{\textbf{WiC}}
    \vspace{-0.5em}

    \begin{tabular*}{\textwidth}{@{\extracolsep{\fill}}l ccccc r}
    \toprule    
     & \textbf{Methods} & 1e-5 & 1e-4 & 1e-3 & 1e-2 & 1e-1 \\ \hline
    
    \midrule

     & \textbf{Sensitive (C4, static)} ($\eps=$1e-3) & 1e-5 & 2e-6 & 1e-6 & 1e-6 & 1e-6 \\ 

     & \textbf{Sensitive (task-specific, static)} ($\eps=$1e-3) & 1e-5 & 2e-6 & 1e-6 & 1e-6 & 1e-6 \\ 

     & \textbf{Sensitive (task-specific, dynamic)} ($\eps=$1e-3) & 1e-5 & 2e-6 & 1e-6 & 1e-6 & 1e-6 \\ 

     & Random (static) ($\eps=$1e-3) & 2e-2 & 5e-3 & 5e-4 & 5e-5 & 5e-6 \\ 

     & Random (dynamic) ($\eps=$1e-3) & 2e-2 & 5e-3 & 5e-4 & 5e-5 & 5e-6 \\ 

     & Weight outliers (static) ($\eps=$1e-3) & 1e-3 & 5e-4 & 2e-4 & 1e-4 & 2e-5 \\ 

    \bottomrule 
    \end{tabular*}
    \end{subtable}

    \begin{subtable}[!ht]{\textwidth}
    \vspace{0.2em}
    \caption{\textbf{COPA}}
    \vspace{-0.5em}

    \begin{tabular*}{\textwidth}{@{\extracolsep{\fill}}l ccccc r}
    \toprule    
     & \textbf{Methods} & 1e-5 & 1e-4 & 1e-3 & 1e-2 & 1e-1 \\ \hline
    
    \midrule

     & \textbf{Sensitive (C4, static)} ($\eps=$1e-3) & 5e-6 & 1e-6 & 1e-6 & 1e-6 & 5e-7 \\ 

     & \textbf{Sensitive (task-specific, static)} ($\eps=$1e-3) & 5e-6 & 2e-6 & 2e-6 & 1e-6 & 1e-6  \\ 

     & \textbf{Sensitive (task-specific, dynamic)} ($\eps=$1e-3) & 5e-6 & 1e-6 & 1e-6 & 1e-6 & 1e-6 \\ 

     & Random (static) ($\eps=$1e-3) & 1e-2 & 2e-3 & 5e-4 & 5e-5 & 5e-6 \\ 

     & Random (dynamic) ($\eps=$1e-3) & 2e-3 & 1e-3 & 2e-4 & 2e-5 & 2e-6 \\ 

     & Weight outliers (static) ($\eps=$1e-3) & 1e-3 & 5e-4 & 5e-4 & 1e-4 & 1e-5 \\ 

    \bottomrule 
    \end{tabular*}
    \end{subtable}

\end{table}

Our hyperparameters (learning rate $\eta$, perturbation scaling constant $\eps$, and the number of ICL examples) for Table~\ref{tab:main-results} is reported in Table~\ref{tab:meta-data-main-results} for reproducibility. We use constant $\eta$ and $\eps$ throughout our experiments. We also report the chosen hyperparameter for Figure~\ref{fig:ablation-sensitive-weights-random} and Figure~\ref{fig:ablation-sensitive-static-dynamic} in Table~\ref{tab:meta-data-ablation}. For LoRA, we always add to all linear layers with $r = 8$ and $\alpha = 16$, and for Prefix Tuning, we always add to $W_K$ and $W_V$ with length as 5, as what \citet{malladi2023fine} uses. 

\clearpage
\subsection{Task-Specific Prompts in Experiments}
We describe our task templates in Table~\ref{tab:prompt}.

\input{tables/prompt}

\clearpage
\subsection{Implementation of Sparse Operations in Linear Layers}\label{sec:appendix:sparse_operations}
\input{arxiv/appendix/experiments/sensitive_sparse_system_impl}

\subsection{Hardware, Platform, Libraries, and Other Details for Benchmarking}\label{computational-resources}
Figure~\ref{fig:training-inference-time}, Figure~\ref{fig:speedup}, and Figure~\ref{fig:sparse-add-impl} (subfigure 1 and 3) are trained and evaluated on an internal cluster with 8 Nvidia RTX A6000 GPUs and 2 Intel Xeon Gold 6342 CPUs, with PyTorch version 2.2, HuggingFace version 4.36, and CUDA 12.2. In subfigure 2 and 4 in Figure~\ref{fig:sparse-add-impl}, we use Nvidia A100-SXM4 (40 GiB) and AMD EPYC 7543P 32-Core CPU with PyTorch version 2.1, HuggingFace version 4.38.2, and CUDA 12.2. We use Flash Attention 2 \citep{dao2023flashattention} throughout our experiments, and the base model for benchmarking is always Llama2-7B with Float16 datatype. 

In Figure~\ref{fig:training-inference-time} and Figure~\ref{fig:sparse-add-impl}, we use sequence length of 512 and batch size 16 sampled from WikiText-2 dataset~\citep{merity2016pointer} as a representative computational intensity for ZO training, and for inference we generate 128 tokens with top-$p$ ($p=0.9$) sampling from the prompt ``\textit{Please describe the effect of sparse zeroth-order optimization methods on memory-efficient LLM fine-tuning: }''.

\end{document}

%% file: macros.tex
\usepackage{amsmath}
\usepackage{amsthm}
\usepackage{amsmath}
\usepackage{amssymb}
\usepackage{amsfonts}
\usepackage{mathtools}
\usepackage{amsfonts} 

\usepackage{cleveref}

\newtheorem{theorem}{Theorem}
\newtheorem{definition}{Definition}
\newtheorem{assumption}{Assumption}
\newtheorem{lemma}{Lemma}

\newcommand{\z}{\mathbf{z}}

\newcommand{\w}{\mathbf{w}}
\newcommand{\x}{\mathbf{x}}
\newcommand{\E}{\mathbb{E}}

\newcommand{\zero}{\mathbf{0}}
\newcommand{\R}{\mathbb{R}}
\newcommand{\eps}{\epsilon}
\newcommand{\argmax}{\text{argmax}}
\newcommand{\argmin}{\text{argmin}}

\newcommand{\grad}{\nabla}
\newcommand{\Identity}{\mathbf{I}}
\newcommand{\mask}{\mathbf{m}}

\newcommand{\zograd}{\hat{g}}
\newcommand{\lr}{\eta}
\newcommand{\zbar}{\bar{\z}}
\newcommand{\F}{\mathcal{F}}

\newcommand{\diag}{\text{diag}}
\newcommand{\Data}{\mathcal{D}}

\DeclareMathOperator{\Tr}{Tr}

\newcommand{\val}[2]{$#1_{#2}$}

\definecolor{mydarkred}{rgb}{0.8,0.02,0.02}

\newcommand{\squishlist}{
  \begin{list}{$\bullet$}
    { \setlength{\itemsep}{0pt}      \setlength{\parsep}{3pt}
      \setlength{\topsep}{3pt}       \setlength{\partopsep}{0pt}
      \setlength{\leftmargin}{1.5em} \setlength{\labelwidth}{1em}
      \setlength{\labelsep}{0.5em} } }
\newcommand{\reallysquishlist}{
  \begin{list}{$\bullet$}
    { \setlength{\itemsep}{0pt}    \setlength{\parsep}{0pt}
      \setlength{\topsep}{0pt}     \setlength{\partopsep}{0pt}
      \setlength{\leftmargin}{0.2em} \setlength{\labelwidth}{0.2em}
      \setlength{\labelsep}{0.2em} } }

 \newcommand{\squishend}{
     \end{list} 
 }

\definecolor{tgreen}{HTML}{00FF00}
\definecolor{tdarkyellow}{HTML}{D1A738}
\definecolor{tred}{HTML}{FE0000}

\definecolor{plot0}{HTML}{1F77B4} 
\definecolor{plot1}{HTML}{FF7F0E} 
\definecolor{plot2}{HTML}{2CA02C} 
\definecolor{plot3}{HTML}{D62728} 
\definecolor{plot4}{HTML}{9467BD} 
\definecolor{plot5}{HTML}{8C564B} 
\definecolor{plot6}{HTML}{E377C2} 
\definecolor{plot7}{HTML}{7F7F7F} 
\definecolor{plot8}{HTML}{BCBD22} 
\definecolor{plot9}{HTML}{17BECF}

%% file: arxiv/sections/abstract.tex
\begin{abstract}
    Zeroth-order optimization (ZO) is a memory-efficient strategy for fine-tuning Large Language Models using only forward passes. However, the application of ZO fine-tuning in memory-constrained settings such as mobile phones and laptops is still challenging since full precision forward passes are infeasible. In this study, we address this limitation by integrating sparsity and quantization into ZO fine-tuning of LLMs. Specifically, we investigate the feasibility of fine-tuning an extremely small subset of LLM parameters using ZO. This approach allows the majority of un-tuned parameters to be quantized to accommodate the constraint of limited device memory. Our findings reveal that the pre-training process can identify a set of ``sensitive parameters'' that can guide the ZO fine-tuning of LLMs on downstream tasks. Our results demonstrate that fine-tuning 0.1\% sensitive parameters in the LLM with ZO can outperform the full ZO fine-tuning performance, while offering wall-clock time speedup. Additionally, we show that ZO fine-tuning targeting these 0.1\% sensitive parameters, combined with 4 bit quantization, enables efficient ZO fine-tuning of an Llama2-7B model on a GPU device with less than 8GiB of memory and notably reduced latency.
\end{abstract}

%% file: arxiv/sections/introduction.tex
Large language models (LLMs) have demonstrated superior performance in general-purpose language generation \citep{brown2020language, radford2019language, liu2019roberta}. 
Despite their success, it remains necessary to fine-tune LLMs for specific tasks to achieve optimal results. However, fine-tuning LLMs often requires much more memory compared to the inference process.
Specifically, there are mainly four parts that occupy the memory during fine-tuning LLMs:
\textbf{(1)} the weight parameter itself;
\textbf{(2)} the optimizer state, which contains the information about the past gradient \citep{kingma2014adam};
\textbf{(3)} the weight gradient used to update the parameters;
\textbf{(4)} the activation cached to calculate the weight gradient \citep{liu2024winner};
In previous work like QLoRA \citep{dettmers2023qlora}, it can reduce both \textbf{(1)} and \textbf{(2)} by combining weight quantization and low-rank adaption \citep{lora}, which enables fine-tuning huge LLMs under data-center level GPUs.
However, under more memory-constraint hardware like cell phones, the memory of caching \textbf{(3)} weight gradient and \textbf{(4)} activation required by backpropagation still cannot be overlooked.
The disparity between the demand of LLM fine-tuning and hardware capacity limits the adaptability of LLMs, especially when personalizing them for edge devices.

\textbf{Exploring Zeroth-Order Optimization in LLM Fine-Tuning.} Recently, there has been a resurging interest in zeroth-order (ZO) optimization methods for LLM fine-tuning \citep{malladi2023fine, liu2024sparse, chen2023deepzero}. ZO optimization method perturbs model parameters in random directions and utilize the loss value difference to compute the gradient direction for parameter update. One advantage of ZO methods in LLM fine-tuning is that they do not require backpropagation procedures, which significantly saves the computation and memory. 
In this way, ZO is backpropagation-free and does not need to cache \textbf{(3)} weight gradients and \textbf{(4)} activations during fine-tuning.
In practice, ZO methods have demonstrated the potential to achieve performance comparable to first-order methods in LLM fine-tuning, which opens the doors for various efficient LLM adaptation strategies.

\begin{wrapfigure}{r}{0.42\columnwidth}
  \centering
  \vspace{-1.5em}
  \includegraphics[width=0.42\columnwidth]{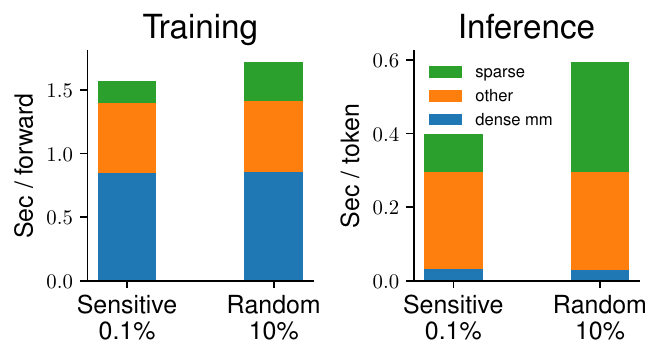}

  \caption{Training \& inference speed of Llama2-7B. As the sensitive sparse fine-tuning method achieves great performance via optimizing only \textit{0.1\%} parameters (performance comparable to ZO full fine-tuning and 10\% random subsets), during inference we achieve an end-to-end $1.49\times$ speedup, with $2.15\times$ speedup at \textcolor{plot2}{sparse operations}.   }
  \label{fig:training-inference-time}
  \vspace{-1.3em}
\end{wrapfigure}

\textbf{Efficient ZO LLM Fine-Tuning with Sparsity.} Although ZO methods remove the need for backpropagation, a significant drawback of these methods is the slow convergence rate  \citep{zhao2024second,liu2024sparse}. A recent approach addresses this by fine-tuning with a sparse mask \citep{liu2024sparse,zhang2024revisiting}, achieving approximately $\sim 75\%$ sparsity. Nonetheless, this sparsity level barely reduces computational overhead, as the latency during the forward pass with \textit{even} $\sim 90\%$ sparsity is still comparable to that of dense matrix operations. This latency increase can greatly impact user experience on applications such as personal assistants, where even a twofold increase in latency is perceptible. In addition, merging the sparse weights back into the base model is impractical on these devices due to memory constraints prohibiting dequantization and quantization. Empirical evidence suggests that higher sparsity levels can significantly decrease the time required for sparse matrix operations, as shown in Figure~\ref{fig:training-inference-time}. This raises the question: 

\emph{Is it possible to leverage the benefits of higher sparsity levels in reducing inference latency while preserving performance on downstream tasks? If so, how far can sparsity be pushed in this context?}

\textbf{Our Proposal: ZO LLM Fine-Tuning with Fisher-Informed, Transferable Sparsity.}
In this paper, we answer the raised research question by proposing an efficient sparse ZO LLM fine-tuning strategy. We observe an extreme sparsity pattern in LLM parameters: a subset, determined by selecting the top $k$ magnitude entries from the empirical Fisher information matrix, is effective for ZO fine-tuning. 
Moreover, we find this sparsity pattern can be obtained through LLM's continuous pre-training process and be transferred to various downstream tasks without modification.

\textbf{Summary of Contributions.} Building on these insights, our work proposes a comprehensive framework for ZO fine-tuning, making the following contributions:

\squishlist

\item We identify that only an extremely small portion (\textbf{0.1\%}) of LLM parameters should be updated during ZO LLM fine-tuning. Moreover, we utilize this insight to guide the memory-efficient on-device personalization of LLMs by low-bit quantization of model parameters.

\item We observe the sparsity pattern observed in LLM pre-training can be transferred across different downstream tasks while still maintaining good ZO performance. Based on this observation, we develop a computational framework to perform parameter-efficient ZO fine-tuning of LLMs.

\item We conduct extensive experiments across various LLMs and demonstrate that our method achieves competitive performance across various downstream tasks.
\squishend

%% file: arxiv/sections/related_works.tex
In this section, we present the formulation for ZO optimization. We also discuss related works about sparsity in LLMs.
\subsection{Zeroth-Order Optimization}\label{sec:method:zo-formulation}

\textbf{ZO surrogate gradient estimator.} ZO optimizers have been studied widely in the machine learning community. Given a dataset $\Data = \{ (\x_1, y_1), \dots, (\x_n, y_n)\}$ and a loss function $f$ with model parameters $\w \in \R^d$, ZO optimizer will estimate the gradient at $\w$ via ZO surrogate gradient estimator. Simultaneous Perturbation Stochastic Approximation (SPSA) \citep{spall1992multivariate} is such an estimator that would first sample a random vector $\z \in \R^d$ and uses the \textit{loss value difference} to scale the update direction. $\z$ is usually sampled from an Gaussian distribution $\mathcal{N}(\zero, \Identity_d)$.

\begin{definition}[\textbf{Simultaneous Perturbation Stochastic Approximation (SPSA)} \citep{spall1992multivariate}] \label{def:SPSA}

    SPSA estimates the gradient w.r.t. $\w$ with a data example $(\x, y)$, a small constant $\eps \in \R$, and a sampled random vector $\z \in \R^d$ as follows:

    \vspace{-1.0em}
    \begin{equation}
        \zograd (\w, (\x, y), \z) = \dfrac{f(\w + \eps \z; (\x, y)) - f(\w - \eps \z; (\x, y))}{2 \eps} \z 
    \end{equation}

\end{definition}

There are other ZO surrogate gradient estimators available \citep{liu2020primer,ohta2020sparse}, but in practice SPSA achieves good performance in ZO optimization, particularly when fine-tuning LLMs. In addition, other ZO algorithms such as DeepZero \citep{chen2023deepzero} would utilize the \textit{parameter-wise} finite difference of loss values to derive \textit{parameter-wise} update directions. This would yield $O(d)$ query costs per training step \textit{even when combining with certain sparse masking methods} and not practical for LLM fine-tuning scenarios. We therefore select SPSA with random Gaussian perturbation as our ZO gradient estimator. 

\textbf{ZO-SGD algorithm.} ZO-SGD is an optimizer similar to SGD but replaces the FO gradient with ZO surrogate gradient estimate per training step, as defined below:
\begin{definition}[\textbf{ZO-SGD update rule}]\label{def:zo_sgd}
    ZO-SGD is an optimizer that uses ZO surrogate gradient to update parameters $\w_t$ with learning rate $\lr_t$ and a data example $(\x_t, y_t)$ sampled at timestep $t$:

    \vspace{-1.0em}
    \begin{equation}
        \w_{t + 1} = \w_t - \lr_t \zograd_\w (\w_t, (\x_t, y_t), \z_t)
    \end{equation}
    \vspace{-1.0em}
    
\end{definition}

MeZO~\cite{malladi2023fine} is a ZO-SGD algorithm that uses the "random seed trick" to save the need of caching ZO surrogate gradient. The choice of optimizer (SGD) is orthogonal to ZO optimization techniques, but in our preliminary experiments we find adaptive optimizers such as Adam \cite{kingma2014adam} would not necessarily accelerate ZO convergence in LLM fine-tuning scenarios. There are other ZO optimizers aware of the parameter-wise heterogeneity of loss curvatures to accelerate the optimization convergence \cite{zhao2024second}, and we leave how to combine our method with theirs as future works. 

\subsection{Sparsity in LLMs} 
Sparsity-driven techniques are widely adopted in improving ML model's efficiency \citep{tan2024sparsity,xia2023flash,liu2023deja,peng2013parallel,frankle2018lottery} and robustness~\cite{zhong2024one,zhong2021revisit}. \citet{frankle2018lottery} showed that within large feed-forward networks, there exists a subnetwork that, when trained in isolation, can achieve test accuracy comparable to that of the original network. In the foundation models era, \citet{liu2023deja} demonstrated that transformer-based models, such as OPT~\cite{opt}, exhibit great sparsity ($\geq 95\%$) in activations. Moreover, \citet{panigrahi2023task} discovered that for RoBERTa \citep{liu2019roberta}, fine-tuning a very small subset of parameters ($\sim 0.01\%$) can yield performance exceeding $95\%$ of that achieved by full fine-tuning.

In the context of ZO optimization, \citet{liu2024sparse} and \citet{zhang2024revisiting} also suggest that sparsity would potentially accelerate ZO optimization convergence. We believe that \textit{ZO has an intrinsic need for sparse training}, as the procedure of ZO gradient estimator usually requires \textit{nearly uniform coordinate-wise scale (in expectation)} perturbation which grows with $d$. In tradition, people usually resolve this with knowledge from parameter-wise loss curvature heterogeneity (replace $\z$ with $\Sigma^{1/2} \z$ where $\Sigma^{1/2}$ serves as a  Hessian-informed preconditioner) \citep{ye2018hessian,zhao2024second}. However, they do not provide a comprehensive investigation on massive parameter models like LLMs. 
In particular, we also observe that during first-order (FO) fine-tuning of LLMs, \textit{the FO gradient can be quite sparse}. We will elaborate more on this insight in the following section (see Figure~\ref{fig:llama2-sparsity} and Figure~\ref{fig:appendix:sparsity}). We would like to explore how sparsity can benefit the ZO LLM fine-tuning.

%% file: arxiv/sections/methods.tex
In this section, we describe the extreme sparsity pattern we observed in LLMs and how we utilize it for efficient ZO fine-tuning including on-device personalization of LLMs.

\subsection{Extreme Sparsity Pattern in LLM}\label{sec:method:sparsity}
\begin{figure*}[!ht]
  \centering

  \includegraphics[width=\textwidth]{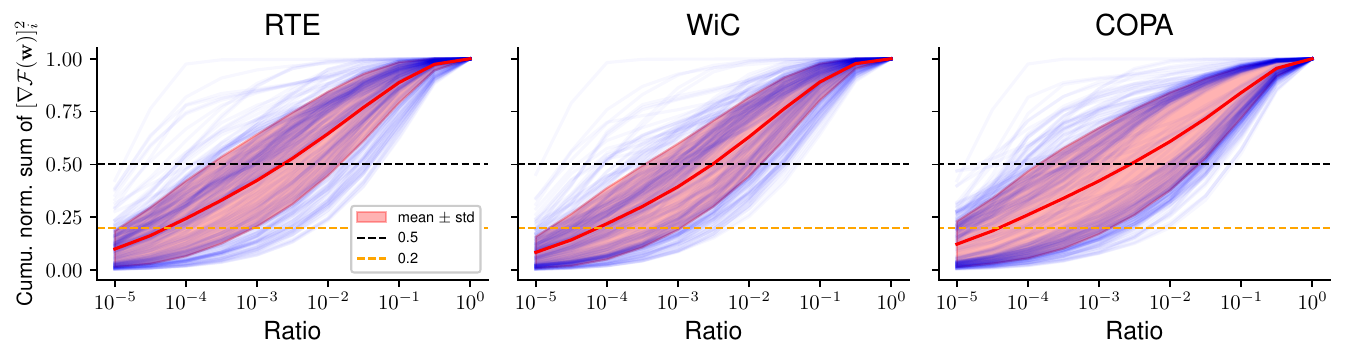}

  \caption{Cumulative normalized sum of coordinate-wise gradient square $[\nabla \mathcal{F}(\w)]_i^2$ of linear layers during Llama2-7B \citep{llama2} fine-tuning. For each linear layer, we first sort parameters by the decreasing order of their gradient square value $[\nabla \mathcal{F}(\w)]_i^2, i \in [d_\text{layer}]$, and we take the cumulative sum and normalize it to draw a \textcolor{blue}{blue curve}, and the \textcolor{red}{red-shaded region} is the mean $\pm$ std of all \textcolor{blue}{blue curves}. More similar figures are in Figure~\ref{fig:appendix:sparsity}.
  \textbf{We observe that roughly 0.1\% parameters in all linear layers contribute about 50\% gradient norm square.}}
  \label{fig:llama2-sparsity}
  \vspace{-.6cm}
\end{figure*}

\textbf{ZO optimization with sensitive parameters.} 
Given model parameters $\w$, a loss function $f$, a data example $(\x, y)$, sensitive parameters are defined as \textit{parameters whose corresponding FO coordinate-wise gradient square values are maximized}. 

\begin{definition}[\textbf{Sensitive parameter mask}\label{def:sensitive}] 

    A sensitive sparse mask $\mask_k \in \{0, 1\}^d$ with $k$ nonzero entries ($\sum_{i} \mask(i) = k$) is defined as
    \begin{equation}
        \mask_k = \argmax_{\mask} \| \mask \odot \nabla f(\w; (\x, y)) \|_2^2.
    \end{equation}
    \vspace{-2em}
\end{definition}

In the context of ZO optimization, we will update sensitive parameters \textit{only}. Denote that $\textcolor{blue}{\zbar} = \textcolor{blue}{\z} \odot \mask_k$. We will modify the SPSA gradient estimator from $\zograd (\w, (\x, y), \textcolor{blue}{\z}) $ to $ \zograd (\w, (\x, y), \textcolor{blue}{\zbar})$, and accordingly:

\begin{definition}[\textbf{Sensitive sparse ZO-SGD update rule}\label{def:sensitive-sparse-sgd}] 

    \begin{equation} 
        \w_{t + 1} = \w_t - \lr_t \zograd_\w (\w_t, (\x_t, y_t), \textcolor{blue}{\zbar_t})
    \end{equation}
    \vspace{-2em}

\end{definition}

The theoretical support of sensitive parameters can be derived from the lens of SPSA gradient estimator and Fisher information matrix as follows:

\reallysquishlist
    \item \textbf{Maximum zeroth-order loss value changes, from the lens of SPSA estimator.}

    The square (account for negativity) of loss value difference for $\zograd_\w (\w_t, (\x_t, y_t), \textcolor{blue}{\zbar_t})$ is as follows:

    \vspace{-1.3em}
    \begin{align}
        \E_{\textcolor{blue}{\zbar}} \{ f(\w + \eps \textcolor{blue}{\zbar}; (\x, y)) - f(\w - \eps \textcolor{blue}{\zbar}; (\x, y)) \}^2 &\approx \E_{\textcolor{blue}{\zbar}} \{2 \eps \textcolor{blue}{\zbar}^\top \nabla_\w f(\w; (\x, y)) \}^2 \\ 
        &= 4 \eps^2 \| \textcolor{blue}{\mask_k \odot}\, \nabla_\w f(\w; (\x, y)) \|^2 
    \end{align}

    Since by Definition~\ref{def:sensitive} our sensitive mask would maximize $\|\mask_k \odot \nabla_\w f(\w; (\x, y)) \|^2 $ for a given sparsity ratio, we would expect our sensitive mask to \textit{maximize} the magnitude of the loss value difference \textit{for any given sparsity ratio}.

    \item \textbf{Maximum coverage of Hessian diagonal, from the lens of Fisher matrix.}

    LLMs are often pre-trained on large text corpus to reach low perplexity before entering the fine-tuning stage. In this case, we would assume $p_\text{LLM}(y | \x) \sim p_\Data(y | \x)$, which implies the empirical Fisher $\hat{\mathbf{F}}$ should be close to the (true) Fisher matrix $\mathbf{F}$ as follows:

    \vspace{-1.3em}
    \begin{align}
        \mathbf{F} &= \E_{\x \sim p_\Data, \hat{y} \sim p_\text{LLM}( \cdot | \x)}\nabla_\w \log p_\text{LLM}(\hat{y} | \x) (\nabla_\w \log p_\text{LLM}(\hat{y} | \x))^\top \\
        &\approx \hat{\mathbf{F}} = \E_{(\x, y) \sim p_\Data}\nabla_\w \log p_\text{LLM}(y | \x) (\nabla_\w \log p_\text{LLM}(y | \x))^\top 
    \end{align}
    \vspace{-1.3em}

    As we assume the empirical Fisher matrix approximates Fisher, which also approximates the Hessian, and empirical Fisher's diagonal is \textit{equal} to the coordinate-wise gradient square vector when computing with downstream task-specific loss, our sensitive parameters would cover a large fraction of the largest Hessian diagonal entries. 
        
\squishend

This idea of sensitive parameters has been studied in the quantization community \cite{kim2023squeezellm,guo2023lq} and FO optimization \citep{sung2021training}. However, \textit{we are the first one to leverage the extremely sparse sensitive parameters in LLM fine-tuning to accelerate ZO fine-tuning with LLMs}. When we have perturbation and updating in the scale of billion parameters, finding which parameters to fine-tune would be important for improving ZO performance. Notice that here we use sensitive masks $\mask_k$ for understanding purposes. In Section~\ref{sec:method:static_sparse}, we will discuss how to transform Definition~\ref{def:sensitive-sparse-sgd} to a parameter-efficient optimization pipeline by optimizing \textit{fixed} sensitive parameters.

\subsection{Theoretical Convergence Rate}\label{sec:method:theory}
We would investigate the theoretical convergence of sensitive sparse ZO-SGD on sensitive parameters under the non-convex optimization settings. Our assumptions are included in Appendix~\ref{appendix:l-smoothness}.

\begin{theorem}[\textbf{Convergence rate of sensitive sparse ZO-SGD (Definition~\ref{def:sensitive-sparse-sgd})}]\label{thm:convergence-rate}

    If we pick $\eta_t = 1 / (L (k + 2))$, under Assumptions~\ref{assumption:bounded-gradient-error} (bounded gradient error), \ref{assumption:l-smooth} (Lipschitz smoothness), and \ref{assumption:sparse_mask} (sparse sensitive parameters), we would have

    \vspace{-1.3em}
    \begin{align}\label{eqn:l-smooth-theory}
        &\dfrac{1}{T} \sum_{t = 0}^{T - 1} \E_{\zbar, (\x,y)} \| \nabla_\w \F(\w_t) \|^2  \leq O\left(\dfrac{k}{c} \cdot \dfrac{L}{T}\right) (\F(\w_0) - \F^*) + 3 \sigma^2.
    \end{align}
    \vspace{-1.3em}

    Moreover, if we still pick $\eta_t = 1 / (L (k + 2))$, with an extra Assumption~\ref{assumption:pl-condition} (P.L. condition), we would have

    \vspace{-1.3em}
    \begin{align}\label{eqn:pl-condition}
         & \E_{\zbar, (\x,y)} \{ \F(\w_T) - \F^* \} \leq  \left(1 - O\left(\dfrac{\mu}{L} \cdot \dfrac{c}{k}\right)\right)^T (\F(\w_0) - \F^*) + \dfrac{3 \sigma^2 c}{2 L (k + 2)}.
    \end{align}
    \vspace{-1.3em}
\end{theorem}

The proof for Inequality~\ref{eqn:l-smooth-theory} is in Appendix~\ref{appendix:l-smoothness} and the proof for Inequality~\ref{eqn:pl-condition} is in Appendix~\ref{appendix:pl-condition}. If we choose $k = d$ and $c = 1$, both convergence rates trivially reduce to the standard zeroth-order convergence rate as $O(d/T) + O(\text{constant})$ and $O((1/d)^T) + O(\text{constant})$. As we assume $c \gg k / d$, we know $d \gg k / c$ and therefore both $O((k/c) (1/T))$ and $O((c/k)^T)$ are much lower than $O(d / T) + O(\text{constant})$ and $O((1/d)^T) + O(\text{constant})$ that zeroth-order method will yield. 

\textit{We want to emphasize that our contributions are more on empirical LLM fine-tuning instead of general machine learning tasks, and in Section~\ref{sec:experiment:effeciveness} we extensively compare our sparse ZO methods with other sparse ZO methods and we demonstrate its superiority during LLM fine-tuning}. We do not use the strict ``local $r$-effective rank'' assumption that \citet{malladi2023fine} uses, and our Assumption~\ref{assumption:sparse_mask} can be easily observed empirically in Figure~\ref{fig:llama2-sparsity}. \citet{liu2024sparse} and \citet{ohta2020sparse} also provide similar analysis on the convergence. However, they do not include our sensitive sparse mask in their studies.

\subsection{Transferability of LLM Pre-Training Sparsity Pattern in ZO Fine-Tuning}\label{sec:method:transferability}
\textbf{Sparse fine-tuning with fixed sensitive parameters.} Our Theorem~\ref{thm:convergence-rate} focuses on \textit{dynamic} sparse fine-tuning. However, \citet{panigrahi2023task} notice that in real LLM fine-tuning scenario, the fine-tuning performance could be attributed to a sparse subset of weights ($\sim 0.01\%$). \citet{malladi2023kernel} also find certain fine-tuning tasks would demonstrate kernel behaviors, which include ``fixed (gradient) features'': $\nabla_\w f(\textcolor{blue}{\w_\text{after FT}}; (\x, y)) \sim \nabla_\w f(\textcolor{blue}{\w_\text{before FT}}; (\x, y))$.

\begin{figure*}[!ht]
  \centering

  \includegraphics[width=\textwidth]{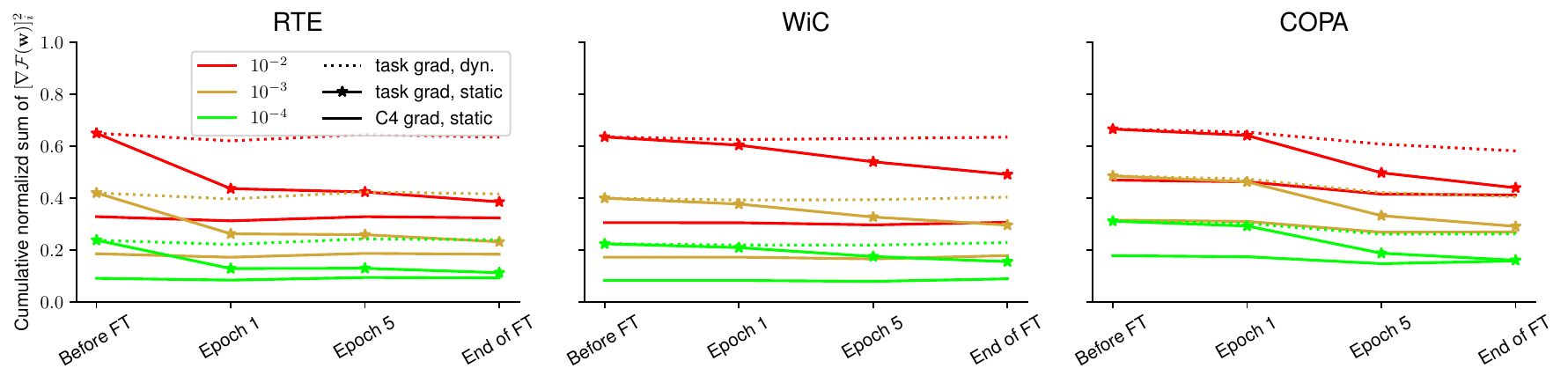}

  \caption{
  Cumulative normalized gradient square values of Llama2-7B model's linear layers during fine-tuning. For each line, the colors represent the fraction of parameters and the line style represents the category. ``task grad, dyn.'' refers to the sensitive parameters selected at the given timestep (x-axis), and ``task grad, static'' refers to the sensitive parameters selected before fine-tuning. ``C4 grad, static'' refers to the sensitive parameters selected with gradients taken from causal language modeling on C4 datasets \citep{2019t5}, and we keep it unchanged during fine-tuning. More similar figures are in Figure~\ref{fig:appendix:transferability}.
  }

  \label{fig:llama2-static}
  \vspace{-.3cm}
\end{figure*}

The similarity of gradient features during fine-tuning  would imply that we do not need to re-select our sensitive parameters during fine-tuning i.e. select once \textit{before fine-tuning} should be sufficient. This hypothesis can be validated by Figure~\ref{fig:llama2-static} and Figure~\ref{fig:ablation-sensitive-static-dynamic}. In Figure~\ref{fig:llama2-static}, the fact that ``task grad, static'' does \textit{not} vanish and still has a large ratio over ``task grad, dyn.'' at the end of training demonstrate that we can select parameters \textit{before fine-tuning}. We also include similar figures for Mistral-7B and OPT-6.7B in Figure~\ref{fig:appendix:transferability} in Appendix~\ref{sec:appendix:transferability}. We will describe Figure~\ref{fig:ablation-sensitive-static-dynamic} in Section~\ref{sec:experiment:on-device-finetuning}.

\textbf{Surrogate sensitive sparse mask from pre-training datasets.} Another observation from Figure~\ref{fig:llama2-static} is that the sensitive parameters derived from pre-training datasets (C4) would still cover a large fraction of model sensitivity. Therefore, we could use it as a \textit{surrogate} sensitive sparse mask when gradients on downstream tasks are unavailable, particularly in scenario of \textit{on-device personalization}. \footnote{Obtaining gradients of LLMs on edge devices is expensive, and we usually cannot transfer data from edge devices to the cloud to compute the gradient on downstream tasks on cloud. In this case we would need some surrogate gradient information to derive sensitive sparse masks on cloud. We will discuss this in Section~\ref{sec:method:on-device-finetuning}.}

\subsection{Our Proposal: ZO LLM Fine-Tuning with Fisher-Informed, Transferable Sparsity}\label{sec:method:static_sparse}
The sparse optimization on \textit{fixed} parameters can be implemented as a parameter-efficient optimization workflow, which will reduce the perturbation and updating time during ZO optimization. Suppose we have derived a sensitive sparse mask $\mask_k$, and we know it is fixed during fine-tuning. Instead of applying $\mask_k$ to $\z$, we would apply it directly to $\w$ and extract the nonzero parts as below:

\vspace{-1.3em}
\begin{equation}\label{eqn:sparse-dense-parameter}
    \textcolor{blue}{\w_\text{sparse}} = \w \odot \mask_k, \quad \w_\text{dense} = \w \odot (\mathbf{1}_d - \mask_k)
\end{equation}
\vspace{-1.3em}

Denote $\z_{k,t} \sim \mathcal{N}(\zero_k, \Identity_k)$ as the Gaussian perturbation sampled in timestep $t$. We will determine $\w_{\text{sparse}}$ before fine-tuning and optimize on $\w_\text{sparse}$ \textit{only} and leave $\w_\text{dense}$ frozen during fine-tuning. In this case, our sensitive sparse ZO-SGD update rule will become:

\vspace{-1.3em}
\begin{equation} \label{eqn:fixed-sensitive-sparse-update-rule}
    \textcolor{blue}{\w_{\text{sparse}, t + 1}} = \textcolor{blue}{\w_{\text{sparse}, t}} - \lr_t \zograd (\textcolor{blue}{\w_{\text{sparse},t}}, (\x_t, y_t), \textcolor{blue}{\z_{k,t}})
\end{equation}
\vspace{-1.3em}

In Section~\ref{sec:method:on-device-finetuning}, we describe how this decomposition would seamlessly combine with existing post-training quantization (PTQ) methods, which creates an opportunity for on-device personalization. In Appendix~\ref{sec:appendix:sparse_operations}, we discuss efficient implementations of linear layers after our decomposition.

\subsection{An Opportunity for On-Device LLM Personalization}\label{sec:method:on-device-finetuning}

\begin{figure*}[!ht]
  \centering

  \includegraphics[width=\textwidth]{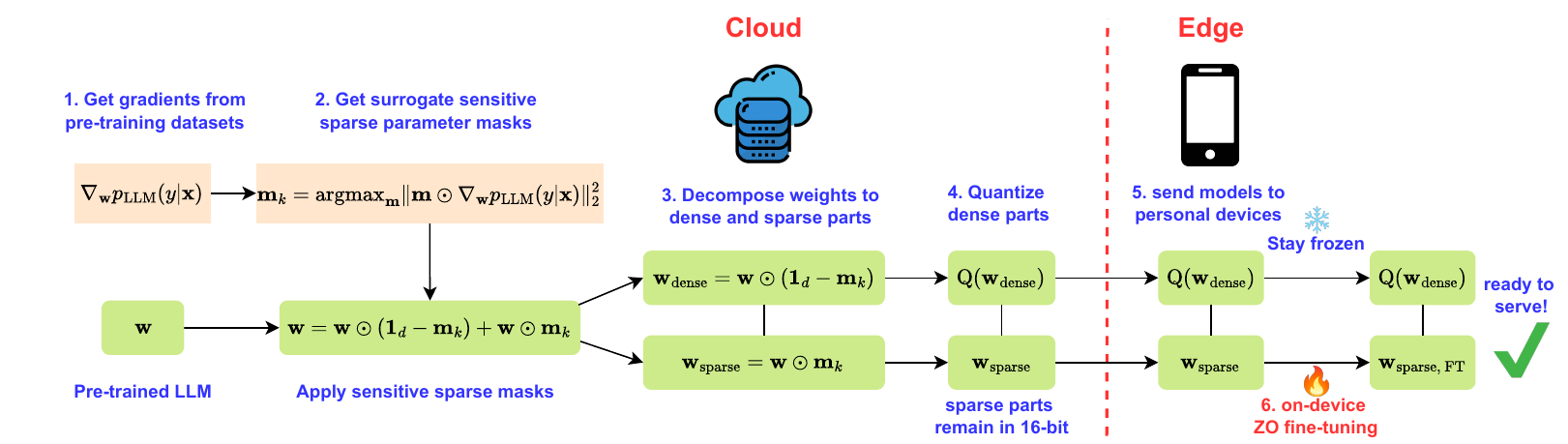}

  \caption{On-device LLM personalization workflow via integrating sensitive sparse ZO optimization with quantization. }

  \label{fig:on-device-workflow}
  \vspace{-1em}
\end{figure*}

As LLMs are often pre-trained with user-agnostic public datasets, personalizing LLMs with individual user's preferences and meet user's specific needs before real-world deployment are vital. \citep{tan2024democratizing,mairittha2020improving} However, transferring the user-specific data to upstream cloud before fine-tuning LLMs would raise privacy concerns. \citep{xu2018deeptype}  On the other hand, personal devices usually have less computational budget and are more memory-constrained than the cloud \citep{zhu2023pockengine}, and performing full fine-tuning would easily exceed the device memory budget.

If we want to fine-tune a 7B-level model (like Llama2-7B) on memory-constrained devices, we need to reduce the memory consumption on \textit{model weights}, \textit{gradients}, \textit{forward activations}, and \textit{optimizer states}:

\squishlist

    \item \textbf{Model weights.}
    We would quantize the $\w_\text{dense}$ to 4 bits, which reduces the model size of a Llama2-7B model from 13.5 to 3.4 GiB.

    \item \textbf{Forward activations.}
    ZO optimization already saves the need of caching activations.

    \item \textbf{Gradients.}
    We would use the ``random seed trick'' same as MeZO \citep{malladi2023fine} to reproduce layer-wise gradients instead of caching them. 

    \item \textbf{Optimizer states.}
    We use SGD. Our method can also be implemented as a parameter-efficient optimization method which is also memory-efficient with other optimizers (even with Adam).

\squishend

As a result, our memory consumption is \textit{nearly minimum}: we can fine-tune a Llama2-7B model under 8 GiB GPU memory \textit{without any offloading}. This would satisfy the memory constraint by a wide range of edge or mobile devices as illustrated in Table~\ref{tab:gpu-memory-spec}.

\textbf{Integration with quantization.}
In Section~\ref{sec:method:static_sparse}, we know that we can obtain surrogate sensitive sparse masks  before fine-tuning. We would first decompose sensitive $\w$ to $\w_{\text{sparse}}$ and $\w_{\text{dense}}$. We will then quantize $\w_{\text{dense}}$. During this process, we will use surrogate gradient information that many PTQ algorithms already have: they need gradients to calibrate their quantization errors. 

Our method also does \textit{not} put strict constraints on specific choices of quantization algorithms since any algorithm \citep{chee2024quip,nagel2020up,frantar2022gptq,lin2023awq,kim2023squeezellm} that aims to minimize the quantization error term or its variant would suffice: 

\vspace{-1.0em}
\begin{equation}
    Q(\w) = \argmin_{Q(\w)} \E_{\x} \| (\w - Q(\w)) \x \|_2^2 
\end{equation}

\textbf{On-device personalization workflow.}
The workflow is illustrated in Figure~\ref{fig:on-device-workflow}. The high-level overview is that we use surrogate gradient information from pre-training datasets $\nabla_\w p_\text{LLM}(y | \x)$ to extract sensitive parameters $\w_\text{sparse}$ and keep $\w_\text{sparse}$ in 16 bits, while we quantize the remaining dense weights $\w_\text{dense}$ (\textcolor{blue}{Step 1-4}). We send $\w_\text{sparse}$ and $Q(\w_\text{dense})$ to personal devices (\textcolor{blue}{Step 5}), and \textbf{we perform on-device ZO fine-tuning only on $\w_\text{sparse}$} (\textcolor{red}{Step 6}).

%% file: arxiv/sections/experiments.tex
In this section, we want to validate the effectiveness of our sensitive sparse ZO optimization method. We also investigate the effectiveness of our on-device personalization recipe in Figure~\ref{fig:on-device-workflow}. There are a few research questions we want to answer:
\squishlist
    \item \textbf{RQ1:} Is optimizing sensitive parameters more effective than optimizing other subset of parameters during ZO fine-tuning? Can we optimize \textit{surrogate} sensitive sparse parameters when downstream gradient information is unavailable?
    
    \item \textbf{RQ2:} Can optimizing extremely sparse and \textit{fixed} parameters (Equation~\ref{eqn:fixed-sensitive-sparse-update-rule}) lead to iteration-wise and total wall-clock time speedup? 
    
    \item \textbf{RQ3:} Can we match the full performance of ZO full fine-tuning by employing our on-device personalization recipe (Figure~\ref{fig:on-device-workflow})?
\squishend

We focus on 7B LLM models (Llama2-7B \citep{llama2}, Mistral-7B \citep{mistral}, OPT-6.7B \citep{opt}) as they would fit with common on-device memory constraints (8 GiB) listed on Table~\ref{tab:gpu-memory-spec} after applying quantization. We use SST-2 
 \citep{sst2}, RTE \citep{glue}, CB \citep{cb}, BoolQ \citep{boolq}, WSC \citep{wsc}, WiC \citep{wic}, and COPA \citep{roemmele2011choice} datasets. We follow standard ZO fine-tuning settings and use the same codebases as in \citet{malladi2023fine}. More details of our experiments (hyperparameters, task-specific prompts, etc.) are in Appendix~\ref{hardware-details}.

\subsection{RQ1: Effectiveness of Sparse ZO Fine-Tuning on Sensitive Parameters}\label{sec:experiment:effeciveness}

We first investigate the performance of optimizing our sensitive parameters versus other subsets of parameters. Our baseline sparsity methods are \textcolor{plot2}{random subsets} and \textcolor{plot1}{weight outliers}. As illustrated in Figure~\ref{fig:ablation-sensitive-weights-random}, we can find that ZO fine-tuning would benefit from sparse optimization, as all methods would achieve higher than ZO full fine-tuning at 90\% sparsity. However, only \textcolor{plot0}{sensitive parameters} would maintain its performance as we move to the extreme sparsity region $(>99\%)$. In fact, \textit{the performance curve of \textcolor{plot0}{sensitive parameters} w.r.t. different sparsity levels is near a flat curve}, which indicates the performance loss by moving from 90\% to 99.9\% is minimal. Therefore, we can optimize \textbf{100 $\times$ less parameters} compared with \textcolor{plot2}{random} and \textcolor{plot1}{weight outliers} and still get same performance.

We also validate whether optimizing \textit{fixed} and \textit{surrogate} sensitive parameters should still yield satisfactory performance. In Figure~\ref{fig:ablation-sensitive-static-dynamic}, we compare the performance of optimizing sensitive parameters with \textcolor{plot0}{C4 gradients} with its theoretical upper bound: \textit{fixed} {sensitive parameters derived from \textcolor{plot4}{task-specific gradients} as the solid line and its dynamic version as the dash-dotted line. We also include the fixed and dynamic \textcolor{plot2}{random subset} parameters as a baseline. We can find that the gap of sensitive parameters between deriving from \textcolor{plot0}{C4 gradients} and \textcolor{plot4}{task-specific gradients} at sparsity level 99.9\% is \textit{small} and \textcolor{plot0}{blue} line is still far above the \textcolor{plot2}{random} and full fine-tuning baseline. We also present a summary of our approaches with 99.9\% sparsity on various datasets and models in Table~\ref{tab:main-results}.

\begin{figure}[!ht]
  \centering
  \begin{minipage}{\linewidth}
    \includegraphics[width=\columnwidth]{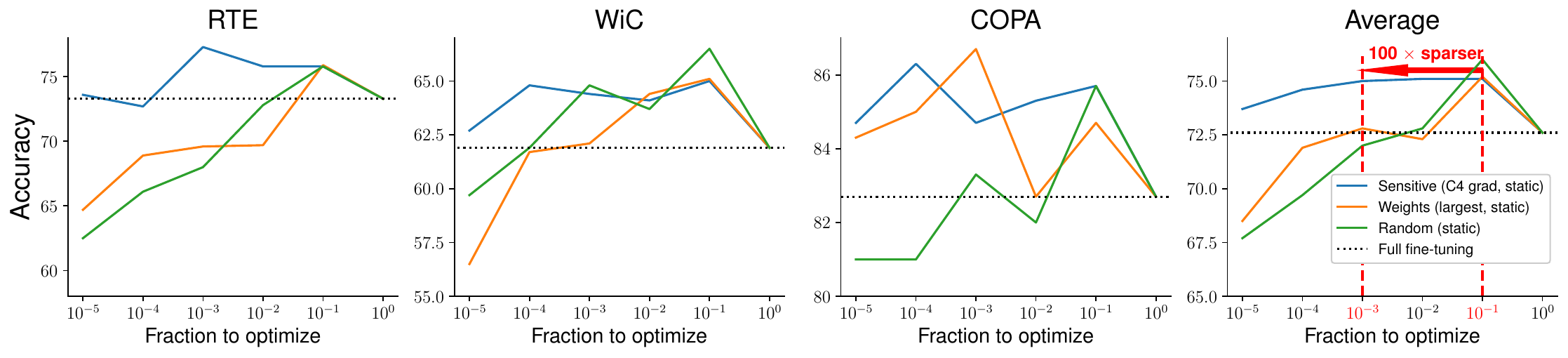}
    \vspace{-1.3em}
    \subcaption{Optimizing \textcolor{plot0}{sensitive parameters with C4 gradients} versus optimizing \textcolor{plot1}{weights with largest magnitude (weight outliers)} and \textcolor{plot2}{random subsets of weights}. The trainable parameters are all determined before fine-tuning and other parameters are kept unchanged.}
    \label{fig:ablation-sensitive-weights-random}  
  \end{minipage}
    \vspace{0.5em}
  \begin{minipage}{\linewidth}
    \includegraphics[width=\columnwidth]{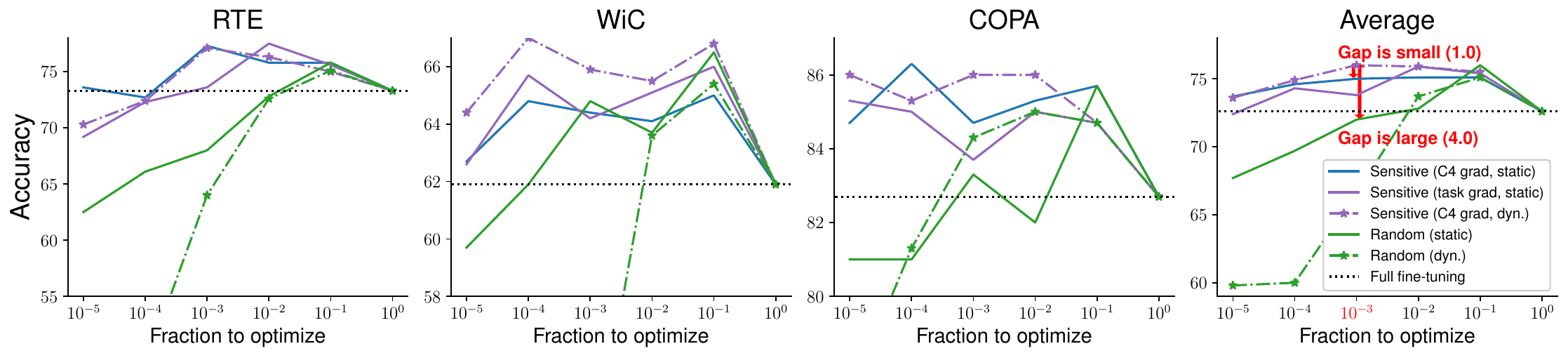}
    \vspace{-1.3em}
    \subcaption{Optimizing sensitive parameters with \textcolor{plot0}{C4 gradients} versus \textcolor{plot4}{task-specific gradients}. ``Static'' means the parameters to optimize are determined before fine-tuning and other parameters are kept unchanged during fine-tuning. ``Dyn.'' means the parameters to optimize will be updated every 100 training steps.}
    \label{fig:ablation-sensitive-static-dynamic}  
  \end{minipage}
  
    \caption{Performance of optimizing sensitive parameters in Llama2-7B fine-tuning on RTE, WiC, and COPA tasks.}
    \label{fig:ablation}
\end{figure}

\begin{figure}[!h]
  \centering

  \vspace{-0.6em}
  \includegraphics[width=\columnwidth]{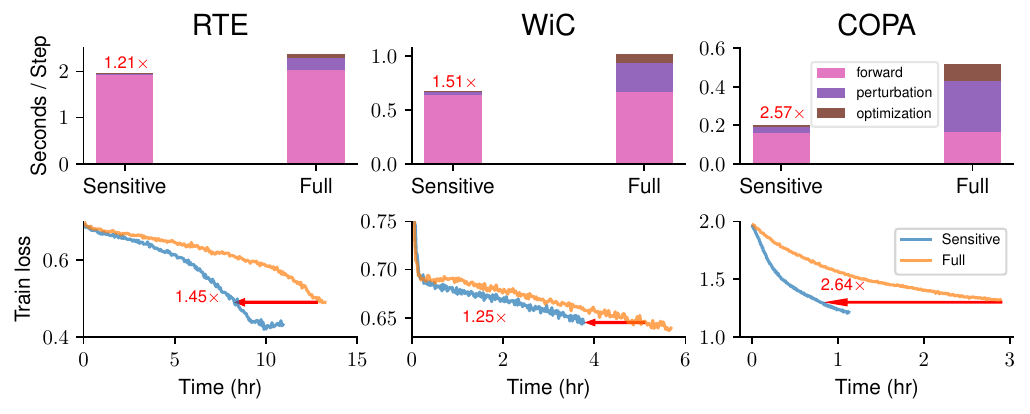}

  \caption{Iteration-wise \& wall-clock convergence time of sensitive sparse fine-tuning on fixed parameters (``\textcolor{plot0}{Sensitive}'') versus ZO full fine-tuning (``\textcolor{plot1}{Full}'') for Llama2-7B. Here we use the 16-bit model as the base model for fine-tuning.  }
  \vspace{-1.3em}
  \label{fig:speedup}
\end{figure}

\begin{table*}[!ht]
     \setlength{\tabcolsep}{3pt}
     \renewcommand{\arraystretch}{1.15} 
     \fontsize{8pt}{8pt}\selectfont
    \centering

    \caption{Performance of difference methods on Llama2-7B fine-tuning tasks. In the first column, ``Q'' means the full model is quantized with 4-bit quantization method (SqueezeLLM~\citep{kim2023squeezellm}), and ``ZO'' means the model is fine-tuned with ZO-SGD optimizer. For each cell, we use the same hyperparameters and repeat it with 3 random seeds. We report the average and standard deviation of test set accuracy in the format of \textbf{\val{\text{mean}}{\text{std}}}. In last 2 columns, ``Acc'' means the average test set accuracy and ``Rank'' means the average rank among all methods across tasks. }  \label{tab:main-results}

    \begin{subtable}[!ht]{\textwidth}
    \caption{\textbf{Llama2-7B}}
    \vspace{-0.5em}

    \begin{tabular*}{\textwidth}{@{\extracolsep{\fill}}l |c | ccccccc | ccr}
    \toprule    
     & \textbf{Methods} & SST-2 & RTE & CB & BoolQ & WSC & WiC & COPA & \textbf{Acc} & \textbf{Rank} \\ \hline
    
    \midrule

     Q, ZO & \textbf{Sensitive (C4, static)} & \val{\textbf{94.7}}{0.4} & \val{\textbf{74.7}}{1.2} & \val{\textbf{66.7}}{2.2} & \val{\textbf{83.0}}{0.5} & \val{57.4}{3.9} & \val{\textbf{65.2}}{0.9} & \val{85.0}{2.2} & \textbf{75.2} & \textbf{2.43} \\ 

     & LoRA & \val{93.8}{0.6} & \val{64.7}{1.1} & \val{64.9}{4.7} & \val{79.7}{1.1} & \val{\textbf{61.5}}{2.1} & \val{59.8}{0.1} & \val{\textbf{85.7}}{0.5} & 72.9 & 4.29 \\ 

     & Prefix & \val{80.5}{4.3} & \val{65.5}{1.2} & \val{63.1}{3.0} & \val{80.3}{0.2} & \val{54.5}{11.4} & \val{58.3}{1.3} & \val{82.0}{0.8} & 69.2 & 5.86 \\ 

    \midrule

    ZO & \textbf{Sensitive (task, static)} & \val{\textbf{94.8}}{0.1} & \val{\textbf{73.6}}{0.9} & \val{\textbf{69.1}}{2.2} & \val{\textbf{83.5}}{0.8} & \val{57.4}{4.7} & \val{64.2}{1.1} & \val{\textbf{83.7}}{2.4} & \textbf{75.2} & \textbf{2.29} \\ 

     & Random (static) & \val{94.1}{0.3} & \val{68.0}{1.7} & \val{64.9}{3.4} & \val{77.0}{0.7} & \val{\textbf{59.6}}{3.6} & \val{\textbf{64.8}}{1.1} & \val{83.3}{1.7} & 73.1 & 4.14  \\ 

     & Full fine-tuning & \val{94.6}{0.5} & \val{73.3}{5.1} & \val{66.7}{0.8} & \val{81.9}{0.8} & \val{58.0}{4.3} & \val{61.9}{0.2} & \val{82.7}{1.7} & 74.2 & 3.57 \\ 

     \midrule

     & Zero-shot & \val{89.0}{0.0} & \val{57.8}{0.0} & \val{32.1}{0.0} & \val{69.9}{0.2} & \val{50.2}{0.0} & \val{36.5}{0.0} & \val{79.0}{0.0} & 59.2 & 7.29 \\

     & ICL & \val{94.8}{0.2} & \val{71.5}{4.3} & \val{72.6}{15.2} & \val{77.5}{4.6} & \val{53.2}{1.1} & \val{61.1}{4.3} & \val{87.0}{2.2} & 74.0 & 3.43 \\

    \bottomrule 
    \end{tabular*}
    
    \end{subtable}

    \begin{subtable}[!ht]{\textwidth}
    \vspace{0.2em}
    \caption{\textbf{Mistral-7B}}
    \vspace{-0.5em}
    \begin{tabular*}{\textwidth}{@{\extracolsep{\fill}}l |c | ccccccc | ccr}

    \midrule

    Q, ZO & \textbf{Sensitive (C4, static)} & \val{\textbf{94.0}}{0.3} & \val{\textbf{74.2}}{2.7} & \val{\textbf{70.2}}{2.2} & \val{\textbf{75.1}}{2.4} & \val{59.6}{4.9} & \val{\textbf{61.2}}{0.9} & \val{\textbf{88.3}}{1.2} & \textbf{74.7} & \textbf{2.86} \\ 

     & LoRA & \val{\textbf{94.0}}{0.4} & \val{65.3}{1.3} & \val{64.9}{4.5} & \val{70.3}{3.7} & \val{\textbf{60.9}}{3.7} & \val{61.1}{0.4} & \val{\textbf{88.3}}{0.5} & 72.1 & 3.57 \\ 

     & Prefix & \val{86.9}{2.1} & \val{57.3}{1.4} & \val{63.7}{5.9} & \val{62.2}{0.9} & \val{60.3}{4.6} & \val{49.0}{0.3} & \val{81.3}{1.7} & 65.8 & 4.86 \\ 

    \midrule

    ZO & \textbf{Sensitive (task, static)} & \val{\textbf{94.7}}{0.3} & \val{\textbf{77.1}}{0.9} & \val{\textbf{69.0}}{0.8} & \val{\textbf{78.4}}{2.2} & \val{\textbf{58.0}}{4.3} & \val{61.4}{0.2} & \val{\textbf{89.3}}{1.3} & \textbf{75.4} & \textbf{1.86} \\ 

     & Random (static) & \val{87.9}{1.9} & \val{50.2}{0.8} & \val{66.1}{4.4} & \val{60.6}{1.7} & \val{57.6}{1.4} & \val{57.3}{0.8} & \val{82.3}{1.7} & 66.0 & 5.29 \\ 

     & Full fine-tuning & \val{94.6}{0.1} & \val{74.6}{2.1} & \val{68.8}{6.2} & \val{76.6}{0.2} & \val{54.8}{6.2} & \val{\textbf{62.6}}{0.5} & \val{88.3}{0.5} & 74.3 & 2.86  \\ 

     \midrule

    & Zero-shot & \val{54.8}{0.0} & \val{50.5}{0.0} & \val{37.5}{0.0} & \val{43.4}{1.8} & \val{50.8}{0.0} & \val{39.4}{0.0} & \val{78.0}{0.0} & 50.6 & 7.00  \\ 

     & ICL & \val{60.7}{16.7} & \val{55.2}{4.7} & \val{33.3}{13.1} & \val{46.8}{6.5} & \val{50.4}{0.6} & \val{63.8}{0.9} & \val{88.7}{0.5} & 57.0 & 5.43  \\ 

    \bottomrule 
    \end{tabular*}   
    
    \end{subtable}

    \begin{subtable}[!ht]{\textwidth}
    \vspace{0.2em}
    \caption{\textbf{OPT-6.7B}}
    \vspace{-0.5em}
    \begin{tabular*}{\textwidth}{@{\extracolsep{\fill}}l |c | ccccccc | ccr}

    \midrule

    Q, ZO & \textbf{Sensitive (C4, static)} & \val{\textbf{94.9}}{0.5} & \val{\textbf{72.8}}{3.6} & \val{\textbf{83.3}}{5.1} & \val{\textbf{73.1}}{0.9} & \val{\textbf{59.3}}{5.3} & \val{\textbf{60.9}}{0.4} & \val{\textbf{84.0}}{1.4} & \textbf{75.5} & \textbf{1.29} \\ 

     & LoRA & \val{94.2}{0.2} & \val{69.6}{1.6} & \val{69.0}{1.7} & \val{69.6}{2.0} & \val{57.1}{9.1} & \val{57.2}{0.8} & \val{83.0}{2.2} & 71.4 & 4.57 \\ 
     
     & Prefix & \val{93.3}{0.4} & \val{71.2}{1.0} & \val{72.0}{1.7} & \val{68.9}{2.8} & \val{62.5}{2.4} & \val{59.4}{0.5} & \val{80.0}{2.4} & 72.5 & 4.14 \\ 

    \midrule

    ZO & \textbf{Sensitive (task, static)} & \val{\textbf{94.5}}{0.4} & \val{\textbf{75.5}}{1.4} & \val{\textbf{82.1}}{3.6} & \val{\textbf{72.5}}{0.8} & \val{\textbf{57.4}}{5.2} & \val{\textbf{60.6}}{1.4} & \val{\textbf{83.3}}{1.7} & \textbf{75.1} & \textbf{2.14} \\ 
     
     & Random (static) & \val{87.3}{2.0} & \val{68.4}{1.7} & \val{70.6}{6.3} & \val{66.0}{1.0} & \val{58.0}{7.0} & \val{56.4}{1.3} & \val{79.0}{0.8} & 69.4 & 5.71 \\ 

     & Full fine-tuning & \val{94.4}{0.3} & \val{72.7}{1.2} & \val{79.8}{3.0} & \val{72.1}{1.2} & \val{\textbf{57.4}}{4.6} & \val{60.2}{0.9} & \val{82.3}{2.6} & 74.1 & 3.29 \\ 

     \midrule

    & Zero-shot & \val{61.0}{0.0} & \val{60.7}{0.0} & \val{46.4}{0.0} & \val{55.7}{1.0} & \val{55.5}{0.0} & \val{36.5}{0.0} & \val{77.0}{0.0} & 56.1 & 7.71 \\ 

     & ICL & \val{74.0}{14.6} & \val{65.8}{11.2} & \val{54.8}{5.9} & \val{67.9}{2.1} & \val{53.2}{1.7} & \val{41.0}{4.5} & \val{80.7}{2.9} & 62.5 & 6.57 \\

    \bottomrule 
    \end{tabular*}   

    \end{subtable}
\vspace{-2mm}
\end{table*}

\subsection{RQ2: Wall-Clock Time Efficiency}\label{sec:experiment:system}
By employing parameter-efficient ZO fine-tuning with extreme sparsity, we also achieve 1.2 - 2.5$\times$ wall-clock time convergence speedup compared with ZO full fine-tuning as we nearly eliminate the \textcolor{plot4}{ZO perturbation} and \textcolor{plot5}{optimizer update} time, as Figure~\ref{fig:speedup} shows. This also boosts the GPU utilization rate as large-batched ZO forward is often compute-bounded while the perturbation and optimization steps are often memory-bounded. Furthermore, the reduced memory footprint of parameter-efficient ZO fine-tuning allows for training larger models on the same hardware, potentially leading to even better performance. As a result, we answer this question that optimizing extremely sparse and fixed parameters leads to substantial iteration-wise and total wall-clock time improvements.

\subsection{RQ3: On-Device Personalization}\label{sec:experiment:on-device-finetuning}

We validate whether our sensitive sparse ZO optimization method would fit with on-device personalization pipeline described in Section~\ref{sec:method:on-device-finetuning} with Table~\ref{tab:main-results}. We follow the exact recipe as described Figure~\ref{fig:on-device-workflow} to report a number as ``Sensitive (C4, static)'', where we only optimize 0.1\% sensitive parameters on top of a 4-bit quantized model. As ZO fine-tuning happens \textit{after} model is quantized, ablating on extracting 0.1\% random subsets of parameters would produce a \textit{different} quantized model. So we choose to report the result for optimizing a fixed random subset on top of the 16-bit model as the ``Random (static)''. 

We also compare with optimizing with LoRA \citep{lora} and Prefix Tuning \citep{prefix} with ZO-SGD optimizer on top of the \textit{same} quantized model. We follow the LoRA $r$ and $\alpha$ and prefix length shown in \citet{malladi2023fine}, and for LoRA, we add it to all linear layers same as where our sensitive parameters are extracted. We find that integrating sensitive sparse ZO optimization with on-device personalization pipelines would still yield good performance exceeding all baselines across models and tasks. Particularly, the performance is higher than ICL, and ZO full fine-tuning in 16 bits. In addition, we have surpassed other ZO-PEFT methods and random sparse ZO fine-tuning methods. This demonstrates the superiority of optimizing sensitive parameters \textit{only} in ZO fine-tuning recipes.
We also notice that optimizing sensitive parameters derived from C4 gradients still produce close results as from task-specific gradients (in average less than 1\% accuracy difference). This indicates optimizing \textit{surrogate} sensitive parameters is still empirically successful.

%% file: arxiv/sections/future_works.tex
We have shown that the sensitive parameters provided by the pre-training process can effectively assist in ZO LLMs fine-tuning. Our experiments suggest that the ZO fine-tuning guided by 0.1\% sensitive parameters in the LLM can even perform better than the full parameter ZO fine-tuning. The experiment results also demonstrate that the quantization of parameters other than sensitive parameters allows us to perform ZO fine-tuning of an LLM on limited memory devices.

%% file: arxiv/appendix/proofs.tex
\subsection{Assumptions}
We start with listing standard assumptions in nonconvex optimization literature:

\input{arxiv/appendix/proofs/assumptions}

\subsection{Proof for Equation~\ref{eqn:l-smooth-theory}, Theorem~\ref{thm:convergence-rate}} \label{appendix:l-smoothness}

\input{arxiv/appendix/proofs/proof_zo_grad_norm}
\input{arxiv/appendix/proofs/proof_lipschitz}

\subsection{Proof for Equation~\ref{eqn:pl-condition}, Theorem~\ref{thm:convergence-rate}} \label{appendix:pl-condition}
\input{arxiv/appendix/proofs/proof_pl_condition}

%% file: arxiv/appendix/proofs/assumptions.tex
\begin{assumption}[\textbf{Bounded stochastic gradient errors}]\label{assumption:bounded-gradient-error}

For any data example $(\x, y) \in \Data$ and for any $\w \in \R^d$, denote the full-batched loss function $\F(\w) = \E_{(\x, y) \in \Data} f(\w; (\x, y))$, we have

    \begin{equation}
        \| \nabla_\w f(\w; (\x, y)) - \nabla_\w \F(\w) \|^2 \leq \sigma^2.
    \end{equation}
    
\end{assumption}

\begin{assumption} [\textbf{Lipschitz smoothness}]\label{assumption:l-smooth}
     We assume that $f(\w, \x)$ is $L$-Lipschitz smooth ($L > 0$): for any $\w, \w' \in \R^d$,

    \begin{equation} 
       \| \grad_\w f(\w; (\x, y)) - \grad_\w f(\w'; (\x, y)) \| \leq L \| \w - \w' \|. 
    \end{equation}
\end{assumption}

\begin{assumption}[\textbf{PL inequality}]\label{assumption:pl-condition}
    We assume that $\F(\w)$ fulfills the Polyak-Lojasiewicz (PL) condition: there exists some $\mu > 0$, for any $\w \in \R^d$

    \begin{equation}
        \dfrac{1}{2} \| \nabla_\w \F(\w) \|^2 \geq \mu (\F(\w) - \F^*), \quad \F^* \text{ is the minimum value } \F^* = \inf_\w \F(\w). 
    \end{equation}
    
\end{assumption}

Inspired by Figure~\ref{fig:appendix:sparsity}, we would assume the sensitive parameters of $\w$ are sparse. 
\begin{assumption}[\textbf{Sensitive parameters are sparse}]\label{assumption:sparse_mask}

We assume at timestep $t$ $\exists \mask_t \in \{0, 1\}^d $ with the number of nonzero entries as $k$, $\exists c \in [0, 1]$ such that  

$$ \| \mask_t \odot \grad_\w f(\w_t; (\x_t, y_t)) \|^2 = c \| \grad_\w f(\w_t; (\x_t, y_t))\|^2. $$ Here we assume $c \gg k/d$. \footnote{From Figure~\ref{fig:appendix:sparsity}, we know that for $c \sim 0.5$, we only need $k/d \sim 0.001$. In this case $k / c \sim 0.002 d$.}

\end{assumption}

%% file: arxiv/appendix/proofs/proof_zo_grad_norm.tex
\begin{lemma}[\textbf{Sparse ZO surrogate gradient covariance}]
    
    \begin{align*}
        & ~\E_{\zbar} \zograd (\w, (\x, y), \zbar) \zograd (\w, (\x, y), y), \zbar)^\top \\
        = & ~\E_{\zbar_i} [\zbar_i \zbar_i^\top \left((\mask_k \odot \grad_\w f(\w; (\x, y))) (\mask_k \odot \grad_\w f(\w; (\x, y)))^\top \right) \zbar_i \zbar_i^\top ] \\
        = & ~2 \left((\mask_k \odot \grad_\w f(\w; (\x, y))) (\mask_k \odot \grad_\w f(\w; (\x, y)))^\top \right) + \| \mask_k \odot \grad_\w f(\w; (\x, y)) \|^2 \tilde{\Identity}_{d, \mask_k} \\
        = & ~2 (\mask_k \mask_k^\top) (\grad_\w f(\w; (\x, y)) \grad_\w f(\w; (\x, y))^\top) + c \| \grad_\w f(\w; (\x, y))\|^2 \tilde{\Identity}_{d, \mask_k} \\
        = & ~2 \tilde{\Identity}_{d, \mask_k} (\grad_\w f(\w; (\x, y)) \grad_\w f(\w; (\x, y))^\top) + c \| \grad_\w f(\w; (\x, y))\|^2 \tilde{\Identity}_{d, \mask_k} 
    \end{align*}
\end{lemma}

%% file: arxiv/appendix/proofs/proof_lipschitz.tex
\begin{proof}[\textbf{Proof for Equation~\ref{eqn:l-smooth-theory}, Theorem~\ref{thm:convergence-rate}}]\label{proof:l-smoothness}
    Denote $\tilde{\Identity}_{d, \mask_k} \in \R^{d \times d}$ as the identity matrix $\Identity_d \in \R^{d \times d}$ with the diagonal masked by $\mask_k \in \{0, 1\}^d$ with $k$ nonzero entries: 
    \begin{align*}
        \diag(\tilde{\Identity}_{d, \mask_k}) &= \mask \odot \mathbf{1}_d, \quad \forall i \neq j, \tilde{\Identity}_{d, \mask_k}(i, j) = \Identity_d(i, j) =  0    
    \end{align*}

    \begin{align*}
        f(\w_{t + 1}, \x_t) &\leq f(\w_t; (\x_t, y_t)) + \langle \nabla f(\w_t; (\x_t, y_t)), \w_{t + 1} - \w_t \rangle + \dfrac{L}{2} \| \w_{t + 1} - \w_t \|^2 \\
        &\leq f(\w_t; (\x_t, y_t)) -  \eta_t \langle \nabla f(\w_t; (\x_t, y_t)), \zograd (\w_t, \x_t, \zbar_t)  \rangle + \dfrac{L \eta_t^2}{2} \| \zograd (\w_t, \x_t, \zbar_t)  \|^2 \\
        \E_{\zbar} f(\w_{t + 1}, \x_t) &\leq \E_{\zbar} f(\w_t; (\x_t, y_t)) -  \eta_t \E_{\zbar} \| \mask_{k, t} \odot \nabla f(\w_t; (\x_t, y_t)) \|^2 + \dfrac{L \eta_t^2}{2} \E_{\zbar} \| \zograd (\w_t, \x_t, \zbar) \|^2 \\
        \E_{\zbar} f(\w_{t + 1}, \x_t) &\leq \E_{\zbar} f(\w_t; (\x_t, y_t)) -  c \eta_t \E_{\zbar} \| \nabla f(\w_t; (\x_t, y_t)) \|^2 + \dfrac{L \eta_t^2}{2} c (k + 2) \E_{\zbar} \| \grad_\w f(\w_t; (\x_t, y_t)) \|^2 \\
        \E_{\zbar,(\x,y)}\F (\w_{t + 1}) &\leq  \E_{\zbar,(\x,y)} \{ \F(\w_t) -  c \eta_t \| \nabla_\w \F(\w_t) \|^2 + c \sigma^2 \eta_t + \dfrac{L \eta_t^2}{2} c (k + 2) \| \grad_\w \F(\w_t) \|^2 + \dfrac{L \eta_t^2}{2} c (k + 2) \sigma^2 \} \\
        \E_{\zbar,(\x,y)} \F(\w_{t + 1}) &\leq \E_{\zbar,(\x,y)} \{\F(\w_t) -  \left(c \eta_t - \dfrac{L \eta_t^2}{2} c (k + 2) \right) \| \nabla_\w \F(\w_t) \|^2 + \left(c \sigma^2 \eta_t + \dfrac{L \eta_t^2}{2} c (k + 2) \sigma^2 \right) \}
    \end{align*}

    Denote $\alpha = L c (k + 2)$, we will have

    \begin{align*}
        \E_{\zbar,(\x,y)} \F(\w_{t + 1}) &\leq \E_{\zbar,(\x,y)} \{ \F(\w_t) -  \eta_t \left(c - \dfrac{\alpha}{2} \eta_t  \right) \| \nabla_\w \F(\w_t) \|^2 + \left(c \sigma^2 \eta_t + \dfrac{\alpha}{2} \sigma^2 \eta_t^2 \right) \}
    \end{align*}

    Set $\eta_t < \dfrac{c}{\alpha} = \dfrac{1}{L(k + 2)}$, we have

    \begin{align*}
        \E_{\zbar,(\x,y)} \F(\w_{t + 1}) &\leq  \E_{\zbar,(\x,y)} \{ \F(\w_t) -  \dfrac{c \eta_t}{2} \| \nabla \F(\w_t) \|^2 + \left(c \sigma^2 \eta_t + \dfrac{\alpha}{2} \sigma^2 \eta_t^2 \right) \} \\
    \end{align*}

    If we apply our sparse ZO update rule recursively for $T$ steps,

    \begin{align*}
        \dfrac{1}{T} \sum_{t = 0}^{T - 1} \E_{\zbar,(\x,y)} \| \nabla_\w \F(\w_t) \|^2  &\leq \dfrac{2 \alpha}{T c^2} (\F(\w_0) - \F^*) + 
        \dfrac{1}{T} \sum_{t = 0}^{T - 1} \dfrac{\left(c \sigma^2 \eta_t + \dfrac{\alpha}{2} \sigma^2 \eta_t^2 \right)}{\dfrac{c \eta_t}{2}}  \\
        &\leq \dfrac{2 \alpha}{T c^2} (\F(\w_0) - \F^*) +  (2 \sigma^2 + \sigma^2)  \\
        &\leq \dfrac{2 L (k + 2)}{c} \dfrac{1}{T} (\F(\w_0) - \F^*) +  3 \sigma^2  \\
        &\leq O\left(\dfrac{k}{c} \cdot \dfrac{L}{T}\right) (\F(\w_0) - \F^*) +  3 \sigma^2 
    \end{align*}
\end{proof}

%% file: arxiv/appendix/proofs/proof_pl_condition.tex
\begin{lemma}[\textbf{Sparse ZO surrogate gradient norm}]

    \begin{align*}
        \E_{\zbar} [\| \zograd (\w_t, \x_t, \zbar_t)   \|^2] &= (2 + k) c \| \grad_\w f(\w, \x) \|^2 
    \end{align*}
\end{lemma}
    
\begin{proof}[\textbf{Proof for Equation~\ref{eqn:pl-condition}, Theorem~\ref{thm:convergence-rate}}]\label{proof:pl-condition}

    Denote $\kappa$ as the condition number $\kappa = \dfrac{\mu}{L}$.

    \begin{align*}
        \E_{\zbar,(\x,y)} \F(\w_{t + 1}) &\leq  \E_{\zbar,(\x,y)} \{ \F(\w_t) -  \dfrac{c \eta_t}{2} \| \nabla \F(\w_t) \|^2 + \left(c \sigma^2 \eta_t + \dfrac{\alpha}{2} \sigma^2 \eta_t^2 \right) \} \\
        &\leq \E_{\zbar,(\x,y)} \{ \F(\w_t) -  c \mu \eta_t (\F(\w_t) - \F^*) + \left(c \sigma^2 \eta_t + \dfrac{\alpha}{2} \sigma^2 \eta_t^2 \right) \} \\
        \E_{\zbar,(\x,y)} \{ \F(\w_{t + 1}) - \F^* \} &\leq  \E_{\zbar,(\x,y)} \{ (\F (\w_t) - \F^*) -  c \mu \eta_t (\F(\w_t) - \F^*) + \left(c \sigma^2 \eta_t + \dfrac{\alpha}{2} \sigma^2 \eta_t^2 \right) \} \\
        \E_{\zbar,(\x,y)} \{ \F(\w_{t + 1}) - \F^* \} &\leq  \E_{\zbar,(\x,y)} \{ (\F (\w_t) - \F^*) -  c \mu \eta_t (\F(\w_t) - \F^*) + \left(c \sigma^2 \eta_t + \dfrac{\alpha}{2} \sigma^2 \eta_t^2 \right) \} 
    \end{align*}

    Plugging in $\eta_t \leq \dfrac{c}{\alpha}$ and applying recursively for $T$ iterations.
    \begin{align*}
        \E_{\zbar, (\x,y)} \{ \F(\w_T) - \F^* \} &\leq (1 - \dfrac{c \kappa}{(k + 2)})^T (\F(\w_0) - \F^*) + \dfrac{3 \sigma^2 c^2}{2 \alpha}  \\
        &\leq (1 - \dfrac{c \kappa}{(k + 2)})^T (\F(\w_0) - \F^*) + \dfrac{3 \sigma^2 c}{2 L (k + 2)} \\
        &\leq \left(1 - O\left(\dfrac{\mu}{L} \cdot \dfrac{c}{k}\right)\right)^T (\F(\w_0) - \F^*) + \dfrac{3 \sigma^2 c}{2 L (k + 2)}
    \end{align*}

\end{proof}

%% file: tables/gpu_memory_spec.tex
\begin{table}[!htbp]
     \setlength{\tabcolsep}{5pt}
     \renewcommand{\arraystretch}{1.25} 
    \fontsize{9pt}{9pt}\selectfont
    \centering

    \caption{Device memory of some mobile devices or consumer-graded GPUs. }
    \label{tab:gpu-memory-spec}
    
    \begin{tabular}{l | c}
    \toprule    
     \textbf{Devices} & Memory\\ \hline
    
    \midrule
    Nvidia GeForce GTX 1080 Ti & 11 GiB \\ \hline

    Nvidia GeForce RTX 3060 Ti & 8 GiB \\ \hline

    Nvidia Jetson TX2 & 8 GiB \\ \hline

    OPPO Find X7 Ultra \citep{li2024transformer} & 12 GiB \\ \hline

    Samsung Galaxy S10 with Mali-G76 GPU \citep{gim2022memory} & 8 GiB \\ \hline
    \bottomrule 
    \end{tabular}    
    
\end{table}

%% file: tables/prompt.tex
\begingroup
\setlength{\tabcolsep}{8pt} 
\renewcommand{\arraystretch}{1.1} 
\begin{table}[!ht]
    \small
     \caption{Task templates for all experiments. On the left column we include the task name and the model name, and on the right column we describe the exact prompt with \textcolor{blue}{answer candidates}.} \label{tab:prompt}

    \begin{center}
    \centering
    \begin{tabular}{p{0.34\linewidth}p{0.01\linewidth}p{0.53\linewidth}}
    \toprule
    \textbf{Task} && \textbf{Prompts} \\
    \midrule
     SST-2 \newline(Llama2-7B) & & \#\#\# Sentence: <text> \#\#\# Sentiment:  \textcolor{blue}{negative/positive} \\\midrule

     SST-2 \newline(Mistral-7B, OPT-6.7B) & & <text> It was \textcolor{blue}{terrible/great} \\\midrule

    RTE \newline(Llama2-7B) & & Suppose "<premise>" Can we infer that "<hypothesis>"? Yes or No? \textcolor{blue}{Yes/No} \\\midrule

    RTE \newline(Mistral-7B, OPT-6.7B) & & <premise>\newline Does this mean that "<hypothesis>" is true? Yes or No?\newline\textcolor{blue}{Yes/No} \\\midrule

    CB \newline(Llama2-7B, Mistral-7B, OPT-6.7B) & & Suppose <premise> Can we infer that "<hypothesis>"? Yes, No, or Maybe?\newline\textcolor{blue}{Yes/No/Maybe} \\\midrule

    BoolQ \newline(Llama2-7B) & & <passage> <question>? \textcolor{blue}{Yes/No} \\\midrule

    BoolQ \newline(Mistral-7B, OPT-6.7B) & & <passage> <question>? \newline\textcolor{blue}{Yes/No} \\\midrule

    WSC \newline(Llama2-7B, Mistral-7B, OPT-6.7B) & & <text>\newline In the previous sentence, does the pronoun "<span2>" refer to <span1>? Yes or No? \newline\textcolor{blue}{Yes/No} \\\midrule

    WiC \newline(Llama2-7B, Mistral-7B, OPT-6.7B) & & Does the word "<word>" have the same meaning in these two sentences? Yes, No?\newline<sent1>\newline<sent2>\newline\textcolor{blue}{Yes/No} \\\midrule

    COPA \newline(Llama2-7B, Mistral-7B, OPT-6.7B) & & <premise> so/because <candidate> \\\midrule

    \bottomrule
    \end{tabular}
    \end{center}
\end{table}

%% file: arxiv/appendix/experiments/sensitive_sparse_system_impl.tex
Linear layers in LLMs often contribute most parameters \citep{kaplan2020scaling}. Since from Equation~\ref{eqn:sparse-dense-parameter} we know 

\begin{equation}
    \textcolor{blue}{\w_\text{sparse}} = \w \odot \mask_k, \quad \w_\text{dense} = \w \odot (\mathbf{1}_d - \mask_k), \quad \w = \textcolor{blue}{\w_\text{sparse}} + \w_\text{dense}
\end{equation}

and since $\w_\text{dense}$ would have same shape (and same computational intensities) as $\w$, we need to improve \textit{wall-clock time} efficiency of $\w_\text{sparse} \x$ to improve the computational efficiency of linear layers after extracting the sparse parameters. In this case, we would have two different methods to implement the forward pass of linear layers (with \textcolor{red}{induced sparse operation colored in red}):

\begin{align}\
    \w \x &= \w_\text{dense} \x \textcolor{red}{+ \w_\text{sparse} \x} \\
    &= \textcolor{red}{\mathrm{SparseAddMM}}(\mathrm{DenseMM}(\w_\text{dense}, \x), \textcolor{blue}{\w_\text{sparse}}, \x) \label{eqn:uniform-quant} &\text{faster with token generation} \\
    &= (\w_\text{dense} \textcolor{red}{+ \w_\text{sparse}}) \x \\
    &= \mathrm{DenseMM}(\textcolor{red}{\mathrm{SparseAdd}}(\textcolor{blue}{\w_\text{sparse}}, \w_\text{dense}), \x)  &\text{faster with ZO training} \label{eqn:non-uniform-quant}
\end{align}

\begin{figure*}[!ht]
  \centering
  \includegraphics[width=\textwidth]{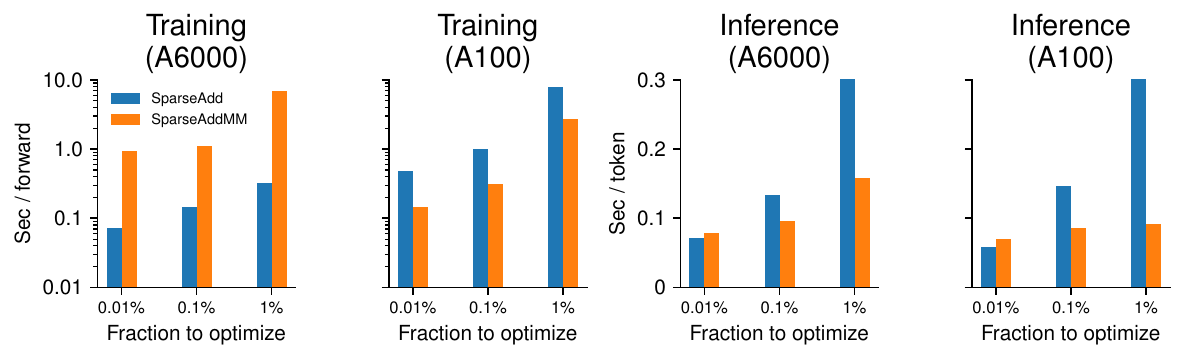}

  \caption{Time of \textcolor{plot0}{SparseAdd} (Equation~\ref{eqn:non-uniform-quant}) versus \textcolor{plot1}{SparseAddMM} (Equation~\ref{eqn:uniform-quant}) in Llama2-7B ZO training forward \& inference. In subfigure 1 and 3, we use Nvidia RTX A6000 and Intel Xeno Gold 6342 CPUs, with PyTorch version 2.2, HuggingFace version 
  4.36, and CUDA 12.2. In subfigure 2 and 4, we use Nvidia A100-SXM4 (40 GiB) and AMD EPYC 7543P 32-Core CPU with PyTorch version 2.1, HuggingFace version 4.38.2, and CUDA 12.2. We use Flash Attention 2 \citep{dao2023flashattention} for all 4 subfigures.}
  \label{fig:sparse-add-impl}
  \vspace{-0.6em}
\end{figure*}

The specific choice of employing Equation~\ref{eqn:uniform-quant} or Equation~\ref{eqn:non-uniform-quant} needs careful consideration and benchmarking, but here we can provide a general guideline based on the size of input vector (or arithmetic intensity) and potential integration with weight quantization method:

\textbf{Size of input vectors $\x$ and arithmetic intensity.}
$\w_\text{sparse} \x$ in Equation~\ref{eqn:uniform-quant} would have a computational dependency over $\x$. During large-batched ZO training, $\x$ would be large enough such that Equation~\ref{eqn:uniform-quant} would induce large computational overhead, as shown in subfigure 1 of Figure~\ref{fig:sparse-add-impl}. In contrast, the computational complexity of Equation~\ref{eqn:non-uniform-quant} is \textit{independent} of $\x$ and when $\x$ is large, we would expect Equation~\ref{eqn:non-uniform-quant} is \textit{much} faster than Equation~\ref{eqn:uniform-quant}. As an example,
we use sequence length of 512 and batch size 16 sampled from WikiText-2 dataset~\citep{merity2016pointer} as a representative computational intensity for ZO training in subfigures 1 and 2 in Figure~\ref{fig:sparse-add-impl}. 

However, during autoregressive token generation, on each step we would only append \textit{a single token} to the previously cached embeddings, and in this case $\x$ is small and computing $\w_\text{dense} + \w_\text{sparse}$ is generally not worthwhile, especially given that $\w_\text{sparse}$ is already sparse. This is also illustrated in subfigure 3 and 4 in Figure~\ref{fig:sparse-add-impl}. However, we note that the specific implementation choice is hardware and task dependent and requires thorough benchmarking and we will leave it as a future work. 

\begin{center}
    \textbf{
    We recommend using Equation~\ref{eqn:non-uniform-quant} during large-batched ZO training and Equation~\ref{eqn:uniform-quant} during small-batched autoregressive token generation.
    }
\end{center}

In light of this observation, in our Figure~\ref{fig:training-inference-time}, we implement both ``SparseAdd'' and ``SparseAddMM'' methods for ``Sensitive (0.1\%)'' and ``Random (10\%)''. For each method we report the \textit{lowest} time out of these 2 implementations: for ``Sensitive (0.1\%)'' training and ``Random (10\%)'' training and inference, we use ``SparseAdd'' approach. For ``Sensitive (0.1\%)'' inference, we use the ``SparseAddMM'' approach.

\textbf{Integration with weight quantization method.}
Weight quantization algorithms can be categorized into 2 categories: uniform quantization method and non-uniform quantization method. For uniform quantization method, \citet{xi2023training} indicates that we could use integer matrix multiplication to compute $Q(\w_\text{dense})\x$ efficiently \textit{without} first dequantizing $Q(\w_\text{dense})$ to 16 bits. However, this creates difficulty on our ``SparseAdd'' approach as we will \textit{violate the constraint of uniformly-spaced quantization bins} by computing $\mathrm{SparseAdd}(Q(\w_\text{dense}) + \w_\text{sparse})$. In this case, we also have 3 different implementations:

\begin{align}\
    Q(\w) \x &\sim Q(\w_\text{dense}) \x \textcolor{red}{+ \w_\text{sparse} \x} \\
    &= \textcolor{red}{\mathrm{SparseAddMM}}\Bigl( \mathrm{Dequantize}\bigl(\mathrm{IntMM}(Q(\w_\text{dense}), \x)\bigr), \textcolor{blue}{\w_\text{sparse}}, \x \Bigr) \label{eqn:quant-1-intmm}  \quad \text{fits with integer matmul }\\
    &= \textcolor{red}{\mathrm{SparseAddMM}}\Bigl( \text{Dequantize}(Q(\w_\text{dense})), \x, \textcolor{blue}{\w_\text{sparse}} \Bigr)  \label{eqn:quant-1-floatmm}\\
    &= (\mathrm{Dequantize}(Q(\w_\text{dense})) \textcolor{red}{+} \textcolor{blue}{\w_\text{sparse}}) \x \\
    &= \mathrm{DenseMM}(\textcolor{red}{\mathrm{SparseAdd}}\left(\textcolor{blue}{\w_\text{sparse}}, \mathrm{Dequantize}(Q(\w_\text{dense})), \x\right) \label{eqn:quant-2}
\end{align}

Equation~\ref{eqn:quant-1-intmm} would compute $\mathrm{IntMM}(Q(\w_\text{dense}), \x)$ \textit{before} dequantizing it to 16 bits. This would make ``SparseAdd'' approach infeasible and we can only employ ``SparseAddMM'' approach in this case. Notice that both Equation~\ref{eqn:quant-1-floatmm} and Equation~\ref{eqn:quant-2} would still dequantize $Q(\w_\text{dense})$ first and the choice of implementation would follow into our discussion of input vector size $\x$ in last paragraph. We will leave a practical implementation and thorough benchmarking into a future work.  

\begin{center}
    \textbf{
    We recommend using Equation~\ref{eqn:quant-1-intmm} when we use efficient integer matmul to compute $Q(\w_\text{dense}) \x$ and in other cases, using Equation~\ref{eqn:quant-1-floatmm} or Equation~\ref{eqn:quant-2} follows our previous recommendation based on the size of input vectors. 
    }
\end{center}

%% file: main.bbl
\begin{thebibliography}{54}
\providecommand{\natexlab}[1]{#1}
\providecommand{\url}[1]{\texttt{#1}}
\expandafter\ifx\csname urlstyle\endcsname\relax
  \providecommand{\doi}[1]{doi: #1}\else
  \providecommand{\doi}{doi: \begingroup \urlstyle{rm}\Url}\fi

\bibitem[Brown et~al.(2020)Brown, Mann, Ryder, Subbiah, Kaplan, Dhariwal, Neelakantan, Shyam, Sastry, Askell, et~al.]{brown2020language}
Tom Brown, Benjamin Mann, Nick Ryder, Melanie Subbiah, Jared~D Kaplan, Prafulla Dhariwal, Arvind Neelakantan, Pranav Shyam, Girish Sastry, Amanda Askell, et~al.
\newblock Language models are few-shot learners.
\newblock \emph{Advances in neural information processing systems}, 33:\penalty0 1877--1901, 2020.

\bibitem[Chee et~al.(2024)Chee, Cai, Kuleshov, and De~Sa]{chee2024quip}
Jerry Chee, Yaohui Cai, Volodymyr Kuleshov, and Christopher~M De~Sa.
\newblock Quip: 2-bit quantization of large language models with guarantees.
\newblock \emph{Advances in Neural Information Processing Systems}, 36, 2024.

\bibitem[Chen et~al.(2024)Chen, Zhang, Jia, Diffenderfer, Liu, Parasyris, Zhang, Zhang, Kailkhura, and Liu]{chen2023deepzero}
Aochuan Chen, Yimeng Zhang, Jinghan Jia, James Diffenderfer, Jiancheng Liu, Konstantinos Parasyris, Yihua Zhang, Zheng Zhang, Bhavya Kailkhura, and Sijia Liu.
\newblock Deepzero: Scaling up zeroth-order optimization for deep model training.
\newblock In \emph{International Conference on Learning Representations}, 2024.
\newblock \doi{10.48550/arXiv.2310.02025}.

\bibitem[Clark et~al.(2019)Clark, Lee, Chang, Kwiatkowski, Collins, and Toutanova]{boolq}
Christopher Clark, Kenton Lee, Ming-Wei Chang, Tom Kwiatkowski, Michael Collins, and Kristina Toutanova.
\newblock Boolq: Exploring the surprising difficulty of natural yes/no questions.
\newblock In \emph{Proceedings of the 2019 Conference of the North American Chapter of the Association for Computational Linguistics: Human Language Technologies, Volume 1 (Long and Short Papers)}, pages 2924--2936, 2019.

\bibitem[Dao(2023)]{dao2023flashattention}
Tri Dao.
\newblock Flashattention-2: Faster attention with better parallelism and work partitioning.
\newblock \emph{arXiv preprint arXiv:2307.08691}, 2023.

\bibitem[De~Marneffe et~al.(2019)De~Marneffe, Simons, and Tonhauser]{cb}
Marie-Catherine De~Marneffe, Mandy Simons, and Judith Tonhauser.
\newblock The commitmentbank: Investigating projection in naturally occurring discourse.
\newblock In \emph{Proceedings of Sinn und Bedeutung}, pages 107--124, 2019.

\bibitem[Dettmers et~al.(2023)Dettmers, Pagnoni, Holtzman, and Zettlemoyer]{dettmers2023qlora}
Tim Dettmers, Artidoro Pagnoni, Ari Holtzman, and Luke Zettlemoyer.
\newblock Qlora: Efficient finetuning of quantized llms.
\newblock In A.~Oh, T.~Naumann, A.~Globerson, K.~Saenko, M.~Hardt, and S.~Levine, editors, \emph{Advances in Neural Information Processing Systems}, volume~36, pages 10088--10115. Curran Associates, Inc., 2023.

\bibitem[Frankle and Carbin(2019)]{frankle2018lottery}
Jonathan Frankle and Michael Carbin.
\newblock The lottery ticket hypothesis: Finding sparse, trainable neural networks.
\newblock In \emph{International Conference on Learning Representations}, 2019.
\newblock \doi{10.48550/arXiv.1803.03635}.

\bibitem[Frantar et~al.(2022)Frantar, Ashkboos, Hoefler, and Alistarh]{frantar2022gptq}
Elias Frantar, Saleh Ashkboos, Torsten Hoefler, and Dan Alistarh.
\newblock Gptq: Accurate post-training quantization for generative pre-trained transformers.
\newblock \emph{arXiv preprint arXiv:2210.17323}, 2022.

\bibitem[Gim and Ko(2022)]{gim2022memory}
In~Gim and JeongGil Ko.
\newblock Memory-efficient dnn training on mobile devices.
\newblock In \emph{Proceedings of the 20th Annual International Conference on Mobile Systems, Applications and Services}, pages 464--476, 2022.

\bibitem[Guo et~al.(2023)Guo, Greengard, Xing, and Kim]{guo2023lq}
Han Guo, Philip Greengard, Eric Xing, and Yoon Kim.
\newblock Lq-lora: Low-rank plus quantized matrix decomposition for efficient language model finetuning.
\newblock In \emph{The Twelfth International Conference on Learning Representations}, 2023.

\bibitem[Hu et~al.(2021)Hu, Shen, Wallis, Allen-Zhu, Li, Wang, Wang, and Chen]{lora}
Edward~J Hu, Yelong Shen, Phillip Wallis, Zeyuan Allen-Zhu, Yuanzhi Li, Shean Wang, Lu~Wang, and Weizhu Chen.
\newblock Lora: Low-rank adaptation of large language models.
\newblock \emph{arXiv preprint arXiv:2106.09685}, 2021.

\bibitem[Jiang et~al.(2023)Jiang, Sablayrolles, Mensch, Bamford, Chaplot, Casas, Bressand, Lengyel, Lample, Saulnier, et~al.]{mistral}
Albert~Q Jiang, Alexandre Sablayrolles, Arthur Mensch, Chris Bamford, Devendra~Singh Chaplot, Diego de~las Casas, Florian Bressand, Gianna Lengyel, Guillaume Lample, Lucile Saulnier, et~al.
\newblock Mistral 7b.
\newblock \emph{arXiv preprint arXiv:2310.06825}, 2023.

\bibitem[Kaplan et~al.(2020)Kaplan, McCandlish, Henighan, Brown, Chess, Child, Gray, Radford, Wu, and Amodei]{kaplan2020scaling}
Jared Kaplan, Sam McCandlish, Tom Henighan, Tom~B Brown, Benjamin Chess, Rewon Child, Scott Gray, Alec Radford, Jeffrey Wu, and Dario Amodei.
\newblock Scaling laws for neural language models.
\newblock \emph{arXiv preprint arXiv:2001.08361}, 2020.

\bibitem[Kim et~al.(2023)Kim, Hooper, Gholami, Dong, Li, Shen, Mahoney, and Keutzer]{kim2023squeezellm}
Sehoon Kim, Coleman Hooper, Amir Gholami, Zhen Dong, Xiuyu Li, Sheng Shen, Michael~W Mahoney, and Kurt Keutzer.
\newblock Squeezellm: Dense-and-sparse quantization.
\newblock \emph{arXiv preprint arXiv:2306.07629}, 2023.

\bibitem[Kingma and Ba(2015)]{kingma2014adam}
Diederik~P Kingma and Jimmy Ba.
\newblock Adam: A method for stochastic optimization.
\newblock In \emph{International Conference on Learning Representations}, 2015.
\newblock \doi{10.48550/arXiv.1412.6980}.

\bibitem[Levesque et~al.(2012)Levesque, Davis, and Morgenstern]{wsc}
Hector Levesque, Ernest Davis, and Leora Morgenstern.
\newblock The winograd schema challenge.
\newblock In \emph{Thirteenth international conference on the principles of knowledge representation and reasoning}, 2012.

\bibitem[Li et~al.(2024)Li, Qian, Lu, Yuan, Wang, and Xie]{li2024transformer}
Luchang Li, Sheng Qian, Jie Lu, Lunxi Yuan, Rui Wang, and Qin Xie.
\newblock Transformer-lite: High-efficiency deployment of large language models on mobile phone gpus.
\newblock \emph{arXiv preprint arXiv:2403.20041}, 2024.

\bibitem[Li and Liang(2021)]{prefix}
Xiang~Lisa Li and Percy Liang.
\newblock Prefix-tuning: Optimizing continuous prompts for generation.
\newblock \emph{arXiv preprint arXiv:2101.00190}, 2021.

\bibitem[Lin et~al.(2023)Lin, Tang, Tang, Yang, Dang, and Han]{lin2023awq}
Ji~Lin, Jiaming Tang, Haotian Tang, Shang Yang, Xingyu Dang, and Song Han.
\newblock Awq: Activation-aware weight quantization for llm compression and acceleration.
\newblock \emph{arXiv preprint arXiv:2306.00978}, 2023.

\bibitem[Liu et~al.(2020)Liu, Chen, Kailkhura, Zhang, Hero~III, and Varshney]{liu2020primer}
Sijia Liu, Pin-Yu Chen, Bhavya Kailkhura, Gaoyuan Zhang, Alfred~O Hero~III, and Pramod~K Varshney.
\newblock A primer on zeroth-order optimization in signal processing and machine learning: Principals, recent advances, and applications.
\newblock \emph{IEEE Signal Processing Magazine}, 37\penalty0 (5):\penalty0 43--54, 2020.

\bibitem[Liu et~al.(2019)Liu, Ott, Goyal, Du, Joshi, Chen, Levy, Lewis, Zettlemoyer, and Stoyanov]{liu2019roberta}
Yinhan Liu, Myle Ott, Naman Goyal, Jingfei Du, Mandar Joshi, Danqi Chen, Omer Levy, Mike Lewis, Luke Zettlemoyer, and Veselin Stoyanov.
\newblock Roberta: A robustly optimized bert pretraining approach.
\newblock \emph{arXiv preprint arXiv:1907.11692}, 2019.

\bibitem[Liu et~al.(2024{\natexlab{a}})Liu, Zhu, Gong, Cheng, Hsieh, and You]{liu2024sparse}
Yong Liu, Zirui Zhu, Chaoyu Gong, Minhao Cheng, Cho-Jui Hsieh, and Yang You.
\newblock Sparse mezo: Less parameters for better performance in zeroth-order llm fine-tuning.
\newblock \emph{arXiv preprint arXiv:2402.15751}, 2024{\natexlab{a}}.

\bibitem[Liu et~al.(2023)Liu, Wang, Dao, Zhou, Yuan, Song, Shrivastava, Zhang, Tian, Re, et~al.]{liu2023deja}
Zichang Liu, Jue Wang, Tri Dao, Tianyi Zhou, Binhang Yuan, Zhao Song, Anshumali Shrivastava, Ce~Zhang, Yuandong Tian, Christopher Re, et~al.
\newblock Deja vu: Contextual sparsity for efficient llms at inference time.
\newblock In \emph{International Conference on Machine Learning}, pages 22137--22176. PMLR, 2023.

\bibitem[Liu et~al.(2024{\natexlab{b}})Liu, Wang, Zhong, Xu, Zha, Tang, Jiang, Zhou, Chaudhary, Xu, et~al.]{liu2024winner}
Zirui Liu, Guanchu Wang, Shaochen~Henry Zhong, Zhaozhuo Xu, Daochen Zha, Ruixiang~Ryan Tang, Zhimeng~Stephen Jiang, Kaixiong Zhou, Vipin Chaudhary, Shuai Xu, et~al.
\newblock Winner-take-all column row sampling for memory efficient adaptation of language model.
\newblock \emph{Advances in Neural Information Processing Systems}, 36, 2024{\natexlab{b}}.

\bibitem[Mairittha et~al.(2020)Mairittha, Mairittha, and Inoue]{mairittha2020improving}
Nattaya Mairittha, Tittaya Mairittha, and Sozo Inoue.
\newblock Improving activity data collection with on-device personalization using fine-tuning.
\newblock In \emph{Adjunct Proceedings of the 2020 ACM International Joint Conference on Pervasive and Ubiquitous Computing and Proceedings of the 2020 ACM International Symposium on Wearable Computers}, pages 255--260, 2020.

\bibitem[Malladi et~al.(2023{\natexlab{a}})Malladi, Gao, Nichani, Damian, Lee, Chen, and Arora]{malladi2023fine}
Sadhika Malladi, Tianyu Gao, Eshaan Nichani, Alex Damian, Jason~D Lee, Danqi Chen, and Sanjeev Arora.
\newblock Fine-tuning language models with just forward passes.
\newblock \emph{Advances in Neural Information Processing Systems}, 36:\penalty0 53038--53075, 2023{\natexlab{a}}.

\bibitem[Malladi et~al.(2023{\natexlab{b}})Malladi, Wettig, Yu, Chen, and Arora]{malladi2023kernel}
Sadhika Malladi, Alexander Wettig, Dingli Yu, Danqi Chen, and Sanjeev Arora.
\newblock A kernel-based view of language model fine-tuning.
\newblock In \emph{International Conference on Machine Learning}, pages 23610--23641. PMLR, 2023{\natexlab{b}}.

\bibitem[Merity et~al.(2016)Merity, Xiong, Bradbury, and Socher]{merity2016pointer}
Stephen Merity, Caiming Xiong, James Bradbury, and Richard Socher.
\newblock Pointer sentinel mixture models.
\newblock In \emph{International Conference on Learning Representations}, 2016.

\bibitem[Nagel et~al.(2020)Nagel, Amjad, Van~Baalen, Louizos, and Blankevoort]{nagel2020up}
Markus Nagel, Rana~Ali Amjad, Mart Van~Baalen, Christos Louizos, and Tijmen Blankevoort.
\newblock Up or down? adaptive rounding for post-training quantization.
\newblock In \emph{International Conference on Machine Learning}, pages 7197--7206. PMLR, 2020.

\bibitem[Ohta et~al.(2020)Ohta, Berger, Sokolov, and Riezler]{ohta2020sparse}
Mayumi Ohta, Nathaniel Berger, Artem Sokolov, and Stefan Riezler.
\newblock Sparse perturbations for improved convergence in stochastic zeroth-order optimization.
\newblock In \emph{Machine Learning, Optimization, and Data Science: 6th International Conference, LOD 2020, Siena, Italy, July 19--23, 2020, Revised Selected Papers, Part II 6}, pages 39--64. Springer, 2020.

\bibitem[Panigrahi et~al.(2023)Panigrahi, Saunshi, Zhao, and Arora]{panigrahi2023task}
Abhishek Panigrahi, Nikunj Saunshi, Haoyu Zhao, and Sanjeev Arora.
\newblock Task-specific skill localization in fine-tuned language models.
\newblock In \emph{International Conference on Machine Learning}, pages 27011--27033. PMLR, 2023.

\bibitem[Peng et~al.(2013)Peng, Yan, and Yin]{peng2013parallel}
Zhimin Peng, Ming Yan, and Wotao Yin.
\newblock Parallel and distributed sparse optimization.
\newblock In \emph{2013 Asilomar conference on signals, systems and computers}, pages 659--646. IEEE, 2013.

\bibitem[Pilehvar and Camacho-Collados(2019)]{wic}
Mohammad~Taher Pilehvar and Jose Camacho-Collados.
\newblock Wic: the word-in-context dataset for evaluating context-sensitive meaning representations.
\newblock In \emph{Proceedings of the 2019 Conference of the North American Chapter of the Association for Computational Linguistics: Human Language Technologies, Volume 1 (Long and Short Papers)}, pages 1267--1273, 2019.

\bibitem[Radford et~al.(2019)Radford, Wu, Child, Luan, Amodei, Sutskever, et~al.]{radford2019language}
Alec Radford, Jeffrey Wu, Rewon Child, David Luan, Dario Amodei, Ilya Sutskever, et~al.
\newblock Language models are unsupervised multitask learners.
\newblock \emph{OpenAI blog}, 1\penalty0 (8):\penalty0 9, 2019.

\bibitem[Raffel et~al.(2019)Raffel, Shazeer, Roberts, Lee, Narang, Matena, Zhou, Li, and Liu]{2019t5}
Colin Raffel, Noam Shazeer, Adam Roberts, Katherine Lee, Sharan Narang, Michael Matena, Yanqi Zhou, Wei Li, and Peter~J. Liu.
\newblock Exploring the limits of transfer learning with a unified text-to-text transformer.
\newblock \emph{arXiv e-prints}, 2019.

\bibitem[Roemmele et~al.(2011)Roemmele, Bejan, and Gordon]{roemmele2011choice}
Melissa Roemmele, Cosmin~Adrian Bejan, and Andrew~S Gordon.
\newblock Choice of plausible alternatives: An evaluation of commonsense causal reasoning.
\newblock In \emph{2011 AAAI Spring Symposium Series}, 2011.

\bibitem[Socher et~al.(2013)Socher, Perelygin, Wu, Chuang, Manning, Ng, and Potts]{sst2}
Richard Socher, Alex Perelygin, Jean Wu, Jason Chuang, Christopher~D Manning, Andrew~Y Ng, and Christopher Potts.
\newblock Recursive deep models for semantic compositionality over a sentiment treebank.
\newblock In \emph{Proceedings of the 2013 conference on empirical methods in natural language processing}, pages 1631--1642, 2013.

\bibitem[Spall(1992)]{spall1992multivariate}
James~C Spall.
\newblock Multivariate stochastic approximation using a simultaneous perturbation gradient approximation.
\newblock \emph{IEEE transactions on automatic control}, 37\penalty0 (3):\penalty0 332--341, 1992.

\bibitem[Sung et~al.(2021)Sung, Nair, and Raffel]{sung2021training}
Yi-Lin Sung, Varun Nair, and Colin~A Raffel.
\newblock Training neural networks with fixed sparse masks.
\newblock \emph{Advances in Neural Information Processing Systems}, 34:\penalty0 24193--24205, 2021.

\bibitem[Tan et~al.(2024{\natexlab{a}})Tan, Zeng, Tian, Liu, Yin, and Jiang]{tan2024democratizing}
Zhaoxuan Tan, Qingkai Zeng, Yijun Tian, Zheyuan Liu, Bing Yin, and Meng Jiang.
\newblock Democratizing large language models via personalized parameter-efficient fine-tuning.
\newblock \emph{arXiv preprint arXiv:2402.04401}, 2024{\natexlab{a}}.

\bibitem[Tan et~al.(2024{\natexlab{b}})Tan, Chen, Zhang, and Liu]{tan2024sparsity}
Zhen Tan, Tianlong Chen, Zhenyu Zhang, and Huan Liu.
\newblock Sparsity-guided holistic explanation for llms with interpretable inference-time intervention.
\newblock In \emph{Proceedings of the AAAI Conference on Artificial Intelligence}, pages 21619--21627, 2024{\natexlab{b}}.

\bibitem[Touvron et~al.(2023)Touvron, Martin, Stone, Albert, Almahairi, Babaei, Bashlykov, Batra, Bhargava, Bhosale, et~al.]{llama2}
Hugo Touvron, Louis Martin, Kevin Stone, Peter Albert, Amjad Almahairi, Yasmine Babaei, Nikolay Bashlykov, Soumya Batra, Prajjwal Bhargava, Shruti Bhosale, et~al.
\newblock Llama 2: Open foundation and fine-tuned chat models.
\newblock \emph{arXiv preprint arXiv:2307.09288}, 2023.

\bibitem[Wang et~al.(2018)Wang, Singh, Michael, Hill, Levy, and Bowman]{glue}
Alex Wang, Amanpreet Singh, Julian Michael, Felix Hill, Omer Levy, and Samuel Bowman.
\newblock Glue: A multi-task benchmark and analysis platform for natural language understanding.
\newblock In \emph{Proceedings of the 2018 EMNLP Workshop BlackboxNLP: Analyzing and Interpreting Neural Networks for NLP}, pages 353--355, 2018.

\bibitem[Xi et~al.(2023)Xi, Li, Chen, and Zhu]{xi2023training}
Haocheng Xi, Changhao Li, Jianfei Chen, and Jun Zhu.
\newblock Training transformers with 4-bit integers.
\newblock \emph{Advances in Neural Information Processing Systems}, 36:\penalty0 49146--49168, 2023.

\bibitem[Xia et~al.(2023)Xia, Zheng, Li, Zhuang, Zhou, Qiu, Li, Lin, and Song]{xia2023flash}
Haojun Xia, Zhen Zheng, Yuchao Li, Donglin Zhuang, Zhongzhu Zhou, Xiafei Qiu, Yong Li, Wei Lin, and Shuaiwen~Leon Song.
\newblock Flash-llm: Enabling cost-effective and highly-efficient large generative model inference with unstructured sparsity.
\newblock In \emph{Proceedings of the VLDB Endowment, Vol. 17, No. 2}, 2023.
\newblock \doi{10.14778/3626292.3626303}.

\bibitem[Xu et~al.(2018)Xu, Qian, Mei, Huang, and Liu]{xu2018deeptype}
Mengwei Xu, Feng Qian, Qiaozhu Mei, Kang Huang, and Xuanzhe Liu.
\newblock Deeptype: On-device deep learning for input personalization service with minimal privacy concern.
\newblock \emph{Proceedings of the ACM on Interactive, Mobile, Wearable and Ubiquitous Technologies}, 2\penalty0 (4):\penalty0 1--26, 2018.

\bibitem[Ye et~al.(2018)Ye, Huang, Fang, Li, and Zhang]{ye2018hessian}
Haishan Ye, Zhichao Huang, Cong Fang, Chris~Junchi Li, and Tong Zhang.
\newblock Hessian-aware zeroth-order optimization for black-box adversarial attack.
\newblock \emph{arXiv preprint arXiv:1812.11377}, 2018.

\bibitem[Zhang et~al.(2022)Zhang, Roller, Goyal, Artetxe, Chen, Chen, Dewan, Diab, Li, Lin, et~al.]{opt}
Susan Zhang, Stephen Roller, Naman Goyal, Mikel Artetxe, Moya Chen, Shuohui Chen, Christopher Dewan, Mona Diab, Xian Li, Xi~Victoria Lin, et~al.
\newblock Opt: Open pre-trained transformer language models.
\newblock \emph{arXiv preprint arXiv:2205.01068}, 2022.

\bibitem[Zhang et~al.(2024)Zhang, Li, Hong, Li, Zhang, Zheng, Chen, Lee, Yin, Hong, et~al.]{zhang2024revisiting}
Yihua Zhang, Pingzhi Li, Junyuan Hong, Jiaxiang Li, Yimeng Zhang, Wenqing Zheng, Pin-Yu Chen, Jason~D Lee, Wotao Yin, Mingyi Hong, et~al.
\newblock Revisiting zeroth-order optimization for memory-efficient llm fine-tuning: A benchmark.
\newblock \emph{arXiv preprint arXiv:2402.11592}, 2024.

\bibitem[Zhao et~al.(2024)Zhao, Dang, Ye, Dai, Qian, and Tsang]{zhao2024second}
Yanjun Zhao, Sizhe Dang, Haishan Ye, Guang Dai, Yi~Qian, and Ivor~W Tsang.
\newblock Second-order fine-tuning without pain for llms: A hessian informed zeroth-order optimizer.
\newblock \emph{arXiv preprint arXiv:2402.15173}, 2024.

\bibitem[Zhong et~al.(2021)Zhong, Zhang, Huang, and Xu]{zhong2021revisit}
Shaochen Zhong, Guanqun Zhang, Ningjia Huang, and Shuai Xu.
\newblock Revisit kernel pruning with lottery regulated grouped convolutions.
\newblock In \emph{International Conference on Learning Representations}, 2021.

\bibitem[Zhong et~al.(2024)Zhong, You, Zhang, Zhao, LeClaire, Liu, Zha, Chaudhary, Xu, and Hu]{zhong2024one}
Shaochen~Henry Zhong, Zaichuan You, Jiamu Zhang, Sebastian Zhao, Zachary LeClaire, Zirui Liu, Daochen Zha, Vipin Chaudhary, Shuai Xu, and Xia Hu.
\newblock One less reason for filter pruning: Gaining free adversarial robustness with structured grouped kernel pruning.
\newblock \emph{Advances in Neural Information Processing Systems}, 36, 2024.

\bibitem[Zhu et~al.(2023)Zhu, Hu, Lin, Chen, Wang, Gan, and Han]{zhu2023pockengine}
Ligeng Zhu, Lanxiang Hu, Ji~Lin, Wei-Ming Chen, Wei-Chen Wang, Chuang Gan, and Song Han.
\newblock Pockengine: Sparse and efficient fine-tuning in a pocket.
\newblock In \emph{Proceedings of the 56th Annual IEEE/ACM International Symposium on Microarchitecture}, pages 1381--1394, 2023.

\end{thebibliography}
